\ifdefined\pdfminorversion
\fi
\documentclass[runningheads]{llncs}
\PassOptionsToPackage{table}{xcolor}

\usepackage{eccv}

\usepackage{eccvabbrv}

\usepackage{array}
\usepackage{graphicx}
\usepackage{booktabs}
\usepackage{makecell}
\usepackage{amsmath, amssymb, amsfonts}
\usepackage{bm}
\usepackage{tabularx}
\usepackage{colortbl}
\usepackage{multirow}
\usepackage[table]{xcolor}
\usepackage{booktabs}
\usepackage{pifont} % Required for the fun checkmarks and crosses!
\usepackage{xcolor}         % Provides extended color capabilities
\usepackage{tabularx}         % Provides extended color capabilities
\usepackage{xkcdcolors}     % Defines XKCD color names for use with `xcolor`.
\newcolumntype{L}[1]{>{\raggedright\arraybackslash}p{#1}}
\newcolumntype{Y}{>{\centering\arraybackslash}X}
\newcolumntype{C}[1]{>{\centering\arraybackslash}p{#1}}
\definecolor{burntorange}{RGB}{204, 85, 0}      % Burnt Orange
\definecolor{crimson}{RGB}{220, 20, 60}         % Crimson
\definecolor{teal}{RGB}{0, 128, 128}            % Teal
\definecolor{royalblue}{RGB}{65, 105, 225}      % Royal Blue
\definecolor{magenta}{RGB}{255, 0, 255}         % Magenta
\definecolor{forestgreen}{RGB}{34, 139, 34}     % Forest Green
\definecolor{deeppink}{RGB}{255, 20, 147}       % Deep Pink
\definecolor{chocolate}{RGB}{210, 105, 30}      % Chocolate
\definecolor{darkviolet}{RGB}{148, 0, 211}      % Dark Violet
\definecolor{cerulean}{RGB}{0, 123, 167}        % Cerulean Blue
\definecolor{olive}{RGB}{128, 128, 0}           % Olive
\usepackage{booktabs}
\usepackage{multirow}
\usepackage[table]{xcolor}
\usepackage{graphicx}

\definecolor{baselinebg}{HTML}{F2F2F2} % Very light gray for baseline
\definecolor{headerbg}{HTML}{E2EFDA}   % Soft green for section header
\definecolor{fluxbg}{HTML}{E2EFDA}     % Soft green to highlight Flux

\definecolor{baselinebg}{HTML}{F2F2F2} % Very light gray for baseline
\definecolor{headerbg}{HTML}{D9EAD3}   % Soft green for section header
\definecolor{initpink}{HTML}{F5CAD9}    % Soft pink for initialized students
\definecolor{oursgreen}{HTML}{F1FAF2}   % Very light green for distilled students
\definecolor{mergegreen}{HTML}{C8F0C8}  % Stronger green for merged checkpoints
\DeclareRobustCommand{\legendbox}[2]{%
    \begingroup\setlength{\fboxsep}{0.35pt}\colorbox{#1}{#2}\endgroup}
\newcommand{\cmark}{\textcolor{green!70!black}{\ding{51}}} % Green check
\newcommand{\xmark}{\textcolor{red!80!black}{\ding{55}}}

\newcommand{\anh}[1]{\textcolor{burntorange}{(Anh Tran: #1)}}

\newcommand{\trung}[1]{\textcolor{darkviolet}{(Trung: #1)}}
\newcommand{\duc}[1]{\textcolor{chocolate}{(Duc: #1)}}
\newcommand{\khoi}[1]{\textcolor{magenta}{(Khoi: #1)}}
\newcommand{\ngan}[1]{\textcolor{cerulean}{(Ngan: #1)}}
\newcommand{\kien}[1]{\textcolor{olive}{(Kien: #1)}}
\newcommand{\quan}[1]{\textcolor{crimson}{(Quan Dao: #1)}}
\newcommand{\viet}[1]{\textcolor{teal}{(Viet: #1)}}
\newcommand{\phong}[1]{\textcolor{royalblue}{(Phong: #1)}}
\newcommand{\cuong}[1]{\textcolor{forestgreen}{(Cuong: #1)}}

\newcommand{\Teacher}{\mathcal{T}}
\newcommand{\Student}{\mathcal{S}}
\newcommand{\Bridge}{\mathcal{B}}
\newcommand{\Loss}{\mathcal{L}}
\newcommand{\LatentT}{\mathcal{Z}_\Teacher}
\newcommand{\LatentS}{\mathcal{Z}_\Student}
\newcommand{\spaceT}{\mathcal{M}_\Teacher}
\newcommand{\spaceS}{\mathcal{M}_\Student}

\newcommand{\mHPSviii}{\textbf{\mbox{HPSv3}}}
\newcommand{\mHPSvii}{\textbf{\mbox{HPSv2}}}
\newcommand{\mIR}{\textbf{\mbox{IR}}}
\newcommand{\mMPS}{\textbf{\mbox{MPS}}}
\newcommand{\mDPG}{\textbf{\mbox{DPG}}}

\newcommand{\myheading}[1]{\vspace{2mm}\noindent{\textbf{#1}}}

\usepackage[most]{tcolorbox}

\newcommand{\finding}[2]{
    \begin{tcolorbox}[
        colback=white!90!gray,
        colframe=teal!60!black,
        arc=5pt,
        boxsep=5pt,
        left=10pt,
        right=10pt,
        top=2pt,
        bottom=2pt,
        boxrule=0.8pt,
        drop shadow=gray!50!white,
        enhanced jigsaw
    ]
    \vspace{-0.1cm}
        \paragraph{\textbf{\textit{}}} #2
    \end{tcolorbox}
    \vspace{-0.1cm}
}

\usepackage{hyperref}

\usepackage{orcidlink}
\usepackage{enumitem}

\hypersetup{
    colorlinks=false,
    citebordercolor=green,
    linkbordercolor={1 1 1},
    urlbordercolor={1 1 1}
}

\begin{document}

\title{Cross-Space Distillation: Teaching One-Step Students with Modern Diffusion Teachers}

\titlerunning{Cross-Space Distillation}

\author{
\href{https://aengusng8.github.io/}{Anh~Nguyen}\inst{1}\textsuperscript{*},
\href{https://scholar.google.com/citations?hl=en&user=QCQPEhoAAAAJ}{Ngan~Nguyen}\inst{1}\textsuperscript{*},
\href{https://danielvu-31.github.io/}{Duc~Vu}\inst{1}\textsuperscript{*}\\
\href{https://trung-dt.com/}{Trung~Dao}\inst{2},
\href{https://viettmab.github.io/}{Viet~Nguyen}\inst{3},
\href{https://quandao10.github.io/}{Quan~Dao}\inst{4}\\
\href{https://scholar.google.com/citations?hl=en&user=FewSoUkAAAAJ}{Kien~Nguyen}\inst{1},
\href{https://scholar.google.com/citations?hl=en&user=D6WZnA0AAAAJ}{Chi~Tran}\inst{1},
\href{https://phongnhhn.info/}{Phong~Nguyen}\inst{1},
\href{https://www.khoinguyen.org/}{Khoi~Nguyen}\inst{1},
\href{https://sites.google.com/view/cuongpham/home/}{Cuong~Pham}\inst{1}\\
\href{https://scholar.google.com/citations?user=a7VNhCIAAAAJ&hl=en}{Dimitris~Metaxas}\inst{4},
\href{https://scholar.google.com/citations?user=AkEXTbIAAAAJ&hl=en}{Vishal~M.~Patel}\inst{3},
and \href{https://scholar.google.com/citations?user=FYZ5ODQAAAAJ&hl=en}{Anh~Tran}\inst{1}
}

\authorrunning{A.~Nguyen et al.}

\institute{%
\begin{tabular}{@{}c@{\qquad}c@{}}
\inst{1}Qualcomm AI Research\textsuperscript{\dag} &
\inst{2}University of Wisconsin--Madison\\
\inst{3}Johns Hopkins University &
\inst{4}Rutgers University
\end{tabular}}

\maketitle
\begingroup
\renewcommand{\thefootnote}{\fnsymbol{footnote}}
\footnotetext[1]{Equal contribution.}
\footnotetext[4]{Qualcomm AI Research is an initiative of Qualcomm Technologies, Inc.}
\endgroup
\vspace{-18pt}
\begin{figure}[!htp]
    \centering
    \includegraphics[width=0.98\linewidth]{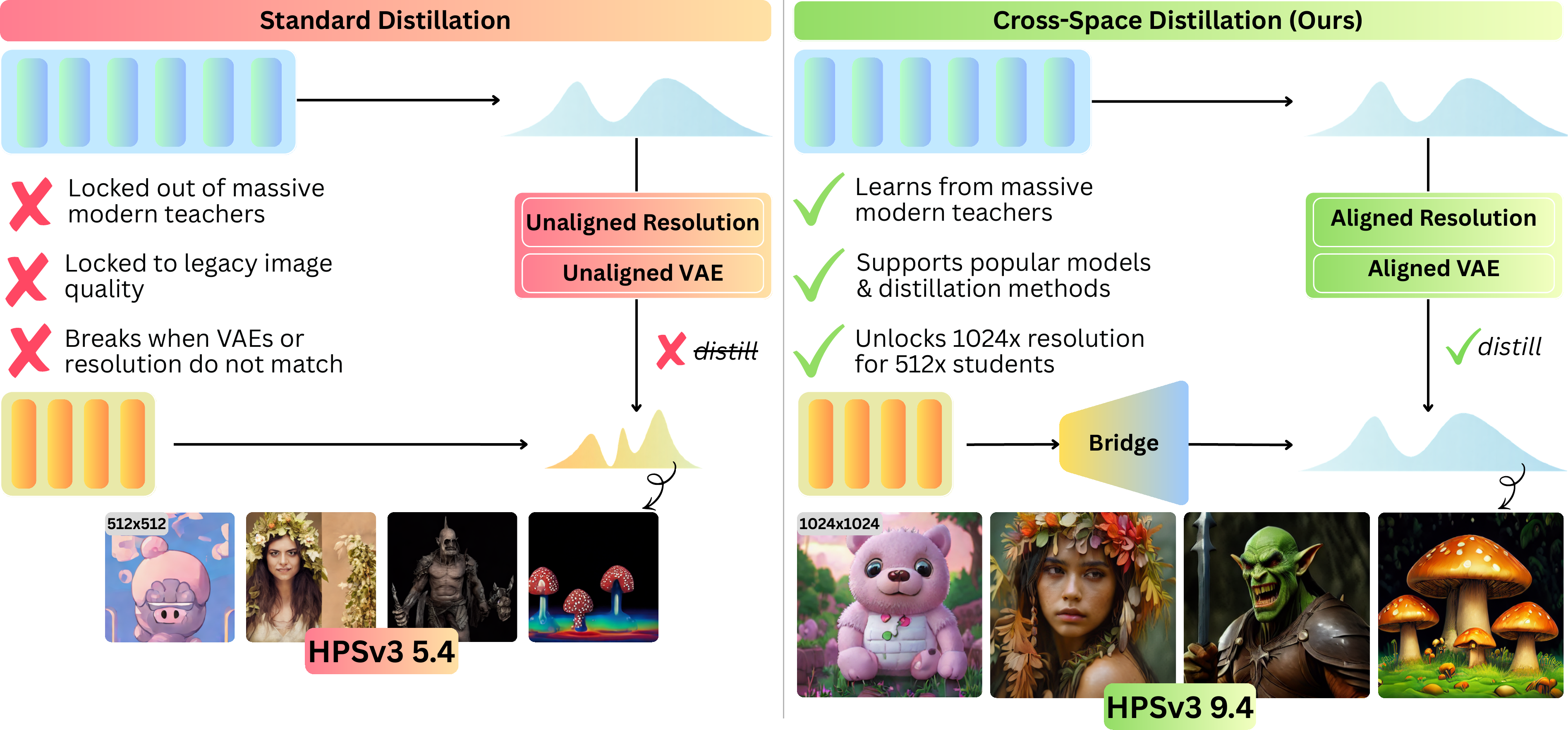}
    \caption{
    \textbf{When Latents Don't Match.} (A) Existing distribution-based distillation methods rely on a \textbf{Shared-Space constraint}, assuming Teacher and Student share the same latent resolution and VAE. This prevents transferring knowledge from high-resolution teachers (\eg, $1024^2$) to compact students (\eg, $512^2$), as their latent tensors are inherently incompatible. (B) We formalize this setting as \textbf{Cross-Space Distillation} and introduce \textbf{Bridge} ($\Bridge_\phi$), a lightweight module that maps student latents $z_S$ to teacher-compatible latents $\hat{z}_T = \Bridge_\phi(z_S)$, enabling standard one-step distillation under Cross-Resolution and Cross-VAE mismatch without modifying the student backbone.
    }
    \vspace{-10mm}
    \label{fig:paradigm}
\end{figure}

\begin{abstract}
Modern one-step diffusion models achieve impressive quality through distribution-based timestep distillation. Yet, they rely on a critical assumption: Teacher and Student must inhabit the same latent space. This Shared-Space constraint prevents knowledge transfer from modern high-capacity Teachers (\eg, SD~3.5 and Flux) into compact, deployment-friendly Students such as SD~1.5, whose latent resolution and VAE parameterization differ from the Teacher. We formalize this overlooked regime as \textbf{Cross-Space Distillation}, where Teacher and Student differ in both latent resolution and VAE space. To enable distillation under this mismatch, we introduce the \textbf{Bridge} ($\Bridge_\phi$), a lightweight latent interface that maps Student latents into the Teacher space without modifying the Student backbone. Bridge combines a frozen Student VAE decoder as a spatial prior with a compact learnable projector, and is trained with latent reconstruction and attention fidelity objectives for stable Teacher-space alignment. Across diverse modern Teachers, Bridge enables substantial gains for compact one-step Students; for example, it improves SD~1.5 from \textbf{5.4} to \textbf{9.4} HPSv3 while preserving one-step inference, low latency, and broad ecosystem compatibility. These results show that heterogeneous large Teachers can be distilled into efficient, deployable backbones through a lightweight latent-space interface.

\keywords{Diffusion Distillation \and One-Step Generation}
\end{abstract}

\section{Introduction}
\label{sec:intro}
% -------- New shorten intro

State-of-the-art text-to-image diffusion models, such as Stable Diffusion 3.5 \cite{sdv3} and Flux \cite{blackforest2024flux1, blackforest2025flux2}, achieve remarkable visual fidelity at high resolutions. However, their large backbones and multi-step sampling impose substantial computational and memory costs, limiting deployment on consumer hardware.

To reduce sampling overhead, recent works distill multi-step teachers into single-step generators via latent consistency distillation \cite{li2024rglcd,zheng2024trajectory,xie2024tlcm}, variational score distillation \cite{sb1,sb2,snoopi}, or adversarial distillation \cite{add,ladd}. Among these approaches, the latter two, which are both grounded in \textit{distribution-based distillation}, have become central to modern one and few-step high-fidelity models. However, they generally preserve the teacher's latent structure and assume a shared latent space, causing the distilled models to inherit large parameter counts or high latent resolutions of the teacher, which limits on-device practicality.

Another line of work compresses large generators through network pruning, structural compression, or lightweight architectures \cite{fastflux,snapfusion,mobilediffusion,snapgen,snapgen++,zhu2025obsdiff}. These methods face three major challenges: pruning often hits a hard sparsity threshold beyond which quality drops, designing compact architectures requires extensive retraining and manual tuning specific to a teacher.

Rather than building new backbones from scratch, we propose reusing compact, widely supported models like Stable Diffusion~1.5 and transferring the advanced generative quality of modern teachers via distillation. This approach is compact, easily integrates with existing ecosystems, is model-agnostic, and benefits from pretrained weights to stabilize and accelerate optimization.

Transferring knowledge to such students introduces four key challenges: \textbf{Cross-Resolution} (from a teacher at 1024$\times$1024 to a student at 512$\times$512), \textbf{Cross-VAE} (different latent spaces), \textbf{Cross-Architecture} (transformers vs. UNets), and \textbf{Cross-Mechanism} (flow matching vs. standard diffusion). Empirically, the architecture and mechanism gaps are comparatively minor; the main barriers are resolution and latent-space mismatch, which prevent standard distillation objectives from being applied directly.

We address these two constraints via \textbf{Cross-Space Distillation}, introducing \textbf{Bridge}, a lightweight module that maps the student latent into the teacher latent space. This enables standard distillation objectives without modifying the student architecture. Bridge uses a compact single-stage design with the student's VAE frozen for spatial alignment, incorporating an Attention Fidelity loss to preserve fine structure and high-quality reconstruction.

\begin{tcolorbox}[
    colback=white!95!gray,
    colframe=teal!60!black,
    arc=4pt,
    boxrule=0.8pt,
    left=7pt, right=7pt, top=6pt, bottom=6pt,
    boxsep=3pt,
    enhanced,
]
\noindent\textbf{Cross-Space Distillation.}
We study distillation where the Teacher and Student do \emph{not} share the same latent space:
\begin{enumerate}[leftmargin=16pt, itemsep=2pt, topsep=5pt, parsep=0pt, partopsep=0pt]
    \item \textbf{Cross-Resolution:} The latent grids have different spatial shapes, \eg,
    $z_S \in \mathbb{R}^{h \times w \times C_S}$ and $z_T \in \mathbb{R}^{H \times W \times C_T}$ with $(h,w)\neq(H,W)$.
    \item \textbf{Cross-VAE:} The autoencoders differ, so latent spaces are not aligned between $\LatentS$ and $\LatentT$.
\end{enumerate}
\noindent\emph{Consequence:} Standard distillation objectives are not directly applicable between $z_S$ and $z_T$ without an explicit alignment mapping.
\end{tcolorbox}

This design preserves backbone compatibility while unlocking high-fidelity outputs. Experiments show that Bridge substantially improves SD~1.5 through Cross-Resolution and Cross-VAE distillation, effectively transferring the teacher's visual priors. Beyond distillation, Bridge is modular, enabling seamless high-resolution synthesis upgrades for existing architectures.

In short, our contributions are as follows:
\begin{enumerate}[leftmargin=16pt, itemsep=2pt, topsep=5pt, parsep=0pt, partopsep=0pt]
    \item \textbf{Cross-Space Distillation}. We formalize a new distillation setting where Teacher and Student differ simultaneously in latent resolution and VAE space, enabling knowledge transfer across mismatched latent spaces.

    \item \textbf{Bridge}. We introduce the Bridge ($\Bridge_\phi$), a lightweight module that maps Student latents into the Teacher space. It combines (i) an architectural prior that leverages the Student's spatial decoder as a scaffold for a compact Teacher-space projector, and (ii) an Attention Fidelity objective preserving structural and fine-grained detail via a reverse-KL on attention distributions.
    \item \textbf{Practical Effectiveness}. Experiments across modern teacher--compact student pairs show substantial improvements in generation quality, pushing compact students close to teacher-level performance while keeping the student backbone unchanged and significantly lighter than the teacher.
\end{enumerate}

\section{Preliminary}
\subsection{Diffusion and Flow}
Diffusion models \cite{ddpm,score_sde,nguyen2024inference} and flow matching models \cite{flow_matching,rectified_flow,cfm_ot,mpdit} provide closely related continuous-time formulations for generative modeling. Both define a forward corruption process that progressively transforms data into Gaussian-like noise and learn a reverse-time generative procedure that maps noise back to the data distribution. During inference, samples are generated by numerically solving the learned reverse dynamics with a discretized solver, typically requiring multiple network evaluations \cite{ddim,dpm_solver,edm}.

\myheading{Diffusion.}
Diffusion models define a fixed forward Gaussian noising process that gradually perturbs clean samples $x_0 \sim p_{\text{data}}$ over $T$ time steps, characterized by
\begin{equation}
x_t = \alpha_t x_0 + \sigma_t \epsilon, \qquad \epsilon \sim \mathcal{N}(0, I),
\end{equation}
where the coefficients $\alpha_t$ and $\sigma_t$ are predefined noise schedules \cite{ddpm}. A neural network, parametrized by $\theta$, is trained to approximate the reverse-time dynamics by gradually removing noise from $x_t$. Common parameterizations of the reverse process include $\epsilon$-prediction, $x_0$-prediction, and $v$-prediction \cite{progressive_distillation}.

\myheading{Flow matching} trains a continuous normalizing flow (CNF) by regressing the conditional velocity field of a chosen probability path \cite{flow_matching}. A common Gaussian probability path uses linear interpolation between data and Gaussian noise,
\begin{equation}
x_t = (1 - t)x_0 + t\epsilon, \qquad t \in (0,1).
\end{equation}
Instead of learning a reverse-time stochastic process, flow matching directly trains a neural network to predict the conditional velocity field $v_\theta(x_t, t)$ that transports samples along this path \cite{flow_matching}. Related formulations include rectified flow \cite{rectified_flow} and conditional/OT flow matching objectives \cite{cfm_ot}.

\myheading{Connection.}
Diffusion and flow matching can be interpreted within a unified continuous-time view where a Gaussian probability path transports data samples to a Gaussian distribution \cite{score_sde,flow_matching}. At endpoints, different parameterizations are related under closed-form transformations at Gaussian endpoints, allowing flow matching velocities to be expressed in terms of diffusion-model outputs \cite{progressive_distillation,diff2flow}. In our work, we convert outputs to a consistent $x_0$-estimate in a shared latent space, enabling Teacher supervision after applying our alignment module.

\subsection{One-Step Distribution-Based Distillation}
\label{section:one_step_distill}
To reduce inference latency to a single network function evaluation, one-step distillation trains a fast generator $\Student_\theta$ to match the sample distribution of a multi-step Teacher \cite{f_distill}. In this work, we focus on \emph{distribution-based} objectives that align the Student distribution with a Teacher-implied target distribution, rather than \emph{trajectory-based} distillation methods that supervise intermediate sampling states or enforce step-to-step consistency, such as consistency distillation \cite{lcm,tcm}.

\myheading{Variational score distillation (VSD)} is a distribution-based recipe that updates the Student by regressing the difference between Student and Teacher scores evaluated on the same noised sample \cite{vsd,f_distill, yin2024onestep, yin2024improved, sb1}.
Let $x=\Student_\theta(\xi,c)$ denote a sample generated by the student model $\Student_\theta$ conditioned on $c$ from Gaussian noise $\xi$. We denote $x_t$ as noised sample of $x$ at time $t$ obtained by the forward diffusion process with $\epsilon\sim\mathcal{N}(0,I)$. The generator update takes the form
\begin{equation}
\nabla_\theta \;\mathbb{E}_{t,\xi,\epsilon}\!\left[\big(s_{\psi}(x_t,t,c)-s_{\eta}(x_t,t,c)\big)\,\nabla_\theta \Student_\theta(\xi,c)\right],
\end{equation}
where $s_{\psi}$ is the frozen Teacher score model and $s_{\eta}$ approximates the intractable Student score.
In practice, $s_{\eta}$ is commonly implemented by an auxiliary score network trained online on Student samples, with alternating updates between $\Student_\theta$ and the score estimator \cite{f_distill}. Besides, this distillation often operates entirely in latent space, assuming the Teacher and Student share the same space.

\myheading{Adversarial distillation.}
Several works introduce a discriminator to align few-step Student outputs with those of a multi-step Teacher by providing an additional distribution-level training signal \cite{ladd,sauer2024adversarial}. This objective is often combined with VSD-style distribution matching to further reduce the gap between a one-step Student and a multi-step Teacher \cite{f_distill,ladd}.

\myheading{In our work.}
These objectives require Teacher and Student to have comparable representations, such as the input to the Teacher score model or any discriminator. Our method satisfies this under Cross-Resolution and Cross-VAE mismatch by aligning Student states into a Teacher-compatible latent space, enabling standard VSD and adversarial objectives with minimal modification.

\section{Distillation Enabled by the Bridge}
\label{sec:method}

\subsection{Cross-Space Distillation}
\label{sec:motivation}
One-step distribution-based distillation methods \cite{sb1, sb2, snoopi, yin2024onestep, yin2024improved, ladd,add} reduce the \emph{temporal} cost of diffusion generation by decreasing the number of sampling steps. However, these methods typically assume that the teacher and student share the same latent space, thereby requiring supervision to be applied directly within that common representation.

This \textbf{Shared-Space constraint} becomes restrictive in practical large-scale distillation scenarios. Modern state-of-the-art teacher models often operate at higher latent resolution and channels and rely on different VAE architectures, whereas compact student models typically employ lower latent resolutions and distinct VAE designs. Under such representational mismatch, standard distillation objectives cannot be directly applied to the teacher and student latent variables without additional alignment.

We refer to this regime as \textbf{Cross-Space Distillation}, in which supervision must be conducted across latent spaces that differ in both \textbf{spatial resolution} and \textbf{underlying VAE representations}. Addressing these discrepancies requires learning an explicit alignment mapping to enable distillation between teacher and student representations.

\subsection{Problem Setting and Notation}

We denote the frozen Teacher denoiser by $\Teacher$ and the Student denoiser by $\Student$.
When referring to autoencoders, we use $(\mathcal{E}_T,\mathcal{D}_T)$ and $(\mathcal{E}_S,\mathcal{D}_S)$ for the Teacher and Student VAE encoder/decoder, respectively.
In this paper, we use the term \emph{space} to refer to a model's \emph{latent representation space} induced by its VAE parameterization and latent grid resolution.
\begin{equation}
\hat{z}_T = \Bridge_\phi(z_S), \qquad \hat{z}_T \in \LatentT.
\end{equation}
A useful Bridge should meet three criteria: (1) High precision in reconstructing the Teacher latent $\hat{z}_T$, (2) Efficiency in memory and computation, adding negligible overhead to the Student network, and (3) Distillation-friendliness, enabling effective knowledge transfer using standard distribution-based distillation methods. To satisfy these requirements, we introduce new designs across Bridge architecture (\cref{sec:bridge_arch}), training objectives (\cref{sec:bridge_objectives}), and describe how the trained Bridge is used for distillation and inference in \cref{sec:bridge_usage}.

\begin{figure}[!t]
    \centering
    \includegraphics[width=1.0\linewidth]{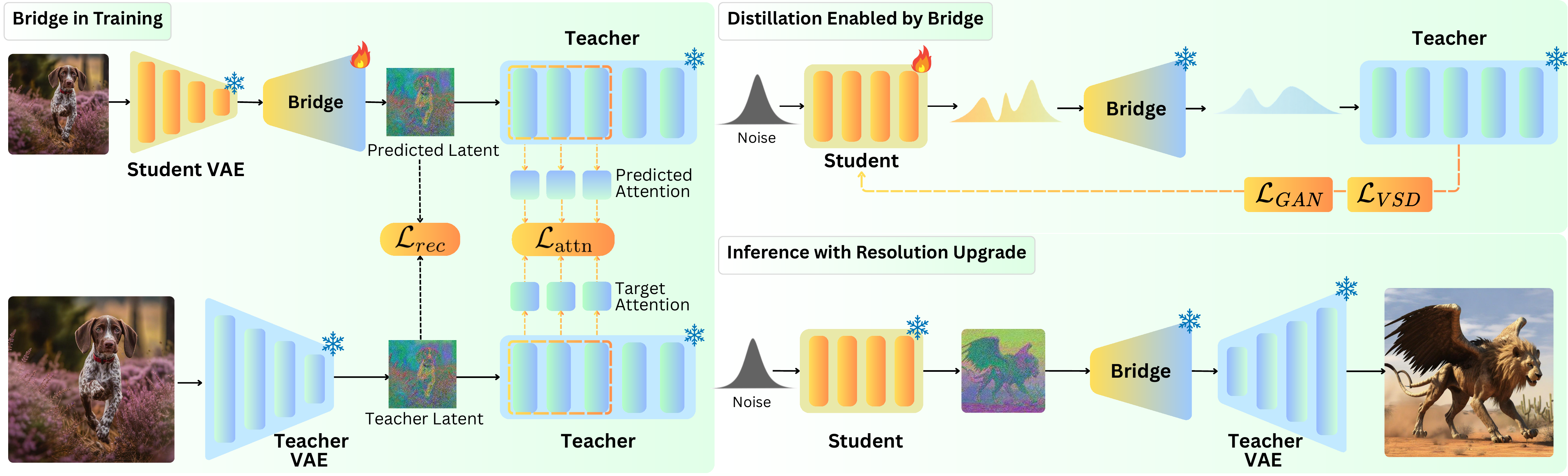}
    \caption{\textbf{Bridge for Cross-Space Distillation.}
    \textbf{Left.} \textbf{Bridge Training:} given an image $x$, we encode it with both VAEs to obtain paired latents $(z_S, z_T)$.
    The Bridge $\mathcal{B}_\phi$ maps $z_S \rightarrow \hat z_T$ using a \textbf{frozen spatial prior}
    from a frozen prefix of the Student decoder $\mathcal{D}_S^{(n)}$ to expand the latent grid,
    followed by a \textbf{learnable projector} $g_\phi$ that outputs a Teacher-compatible latent.
    We train $\mathcal{B}_\phi$ with latent reconstruction $\mathcal{L}_{rec}$ and attention fidelity $\mathcal{L}_{attn}$
    computed from a frozen Teacher denoiser by matching attention responses induced by $\hat z_T$ and $z_T$.
    \textbf{Top-right.} \textbf{Distillation Enabled by Bridge:} after training, $\mathcal{B}_\phi$ converts Student outputs into the Teacher latent space so standard one-step distillation losses such as $\mathcal{L}_{VSD}$ and $\mathcal{L}_{GAN}$ could apply, bypassing Cross-Resolution and Cross-VAE mismatch.
    \textbf{Bottom-right.} \textbf{Inference with Resolution Upgrade:} a low-resolution Student sample can be mapped by $\mathcal{B}_\phi$ and decoded with the Teacher decoder to synthesize at the Teacher resolution without changing the Student denoiser.
    \vspace{-10pt}
    }
    \label{fig:method}
\end{figure}

\subsection{Architecture with Rich Priors}
\label{sec:bridge_arch}

The key challenge in distillation under latent mismatch is that $z_S$ and $z_T$ differ in spatial shape and latent parameterization. We therefore factor $\Bridge_\phi$ into two stages. The first stage aligns spatial resolution from $h \times w$ to $H \times W$. The second stage aligns features into the Teacher latent representation.

\myheading{Spatial prior.}
For spatial alignment, one could learn a latent upsampler jointly with the Bridge, but this requires the model to learn spatial upsampling behavior from scratch. Instead, we reuse the early decoding blocks of the Student VAE decoder, which already implement latent upsampling and are trained with a massive amount of data. Let $\mathcal{D}_{S}^{(n)}$ denote the first $n$ decoding blocks of the Student VAE decoder. We keep them frozen and choose $n$ such that their output has spatial resolution $H \times W$, matching the Teacher's latent resolution. Given $z_S\in\mathbb{R}^{h\times w\times C_S}$, we compute:
\begin{equation}
    f_{\mathrm{prior}} = \mathcal{D}_{S}^{(n)}(z_S),
    \qquad
    f_{\mathrm{prior}} \in \mathbb{R}^{H\times W\times C_{\mathrm{prior}}},
\end{equation}
where $C_{\mathrm{prior}}$ is the channel dimension of this decoder feature. This stage performs the spatial expansion without introducing additional trainable parameters and provides a strong scaffold for the next stage.

\myheading{Projection head.}
After spatial alignment, $f_{\mathrm{prior}}$ has the correct spatial resolution but remains in the Student decoder feature space rather than the Teacher latent representation. We therefore learn a compact projection head $g_\phi$ that maps $f_{\mathrm{prior}}$ to a Teacher compatible latent:
\begin{equation}
    \hat{z}_T = \Bridge_\phi(z_S) = g_\phi(f_{\mathrm{prior}}),
    \qquad
    \hat{z}_T \in \mathbb{R}^{H\times W\times C_T}.
\end{equation}
Overall, our Bridge is:
$
    z_S \xrightarrow{\ \mathcal{D}_{S}^{(n)}\ } f_{\mathrm{prior}} \xrightarrow{\ g_\phi\ } \hat{z}_T,
$
which isolates spatial alignment in the frozen decoder prefix while using the trainable module $g_\phi$ for channel and semantic alignment. We optimize $g_\phi$ with the objectives in \cref{sec:bridge_objectives}.

\subsection{Training Objectives}
\label{sec:bridge_objectives}

We optimize the \emph{Bridge} using two complementary objectives: the first term provides a regression signal in the Teacher latent space, while the second term adds Teacher-based supervision by matching internal attention responses.

\myheading{Latent Reconstruction ($\Loss_{rec}$).}
Given paired latents $(z_S, z_T)$ extracted from the Student and Teacher encoders, the Bridge maps the Student latent to the Teacher latent space as $\hat{z}_T = \Bridge_\phi(z_S)$. We supervise the Bridge with an $\ell_1$ reconstruction loss:
\begin{equation}
    \Loss_{rec} = \| z_T - \hat{z}_T \|_1.
\end{equation}
While $\ell_2$ is a common choice, we observe that it is often inefficient and less effective for latent-space reconstruction in practice, as noted in recent distillation/acceleration works \cite{sdxl_lightning,hipa,improved_shortcut}. In our setting, the $\ell_1$ objective provides better training stability and helps preserve salient structure in the Teacher latent space.

\myheading{Attention Fidelity ($\Loss_{attn}$).}
\label{sec:attn_loss}
While latent reconstruction $\Loss_{rec}$ is necessary for alignment, it is often \textit{insufficient}: small latent discrepancies can produce perceptible semantic errors after decoding. Pixel-wise losses may tolerate slight misalignments, thus losing \textit{fine-grained structure}. Adding adversarial losses can sharpen details but is sensitive to hyperparameters and reduces training stability. We therefore seek a supervision signal that is \textbf{(i)} stable, \textbf{(ii)} globally aware, and \textbf{(iii)} defined directly in Teacher space.

To this end, inspired by \cite{Ma_2025_ICCV}, we align the Teacher denoiser's self-attention \emph{distributions}. Attention captures long-range dependencies, so mismatches between $\hat z_T$ and $z_T$ are amplified in the Teacher's internal responses; a compatible latent should induce similar attention patterns under identical diffusion conditions. We extract attention maps $P^l(z;t,c) \in [0,1]^{N\times N}$ from a fixed set of Teacher layers $l \in \text{Layers}$, where $N$ is the number of tokens. Each $P^l$ is obtained via row-wise softmax over key positions from the Teacher's query and key activations. Following MiniLLM \cite{gu2023minillm}, we use reverse KL divergence, which focuses on the dominant attention mass and is empirically more stable than forward KL. We minimize:
\begin{equation}
\Loss_{attn}
=\sum_{l\in\text{Layers}} \tau^2\,
\mathrm{KL}\!\left(
P^l(\hat{z}_T;t,c)
\,\middle\|\,
P^l(z_T;t,c)
\right),
\end{equation}
where $\tau>0$ is a temperature for numerical stability, and we normalize attention logits before applying the softmax.

\myheading{Final Objective ($\Loss_{final}$).}
We train the Bridge by minimizing a weighted sum of latent reconstruction and attention fidelity:
\begin{equation}
\Loss_{final}=\alpha \Loss_{rec}+\beta \Loss_{attn},
\label{eq:bridge_loss}
\end{equation}
where $\alpha,\beta$ are hyperparameters and set to $\alpha=\beta=1$ in our experiments.

\myheading{Bridge Usage.}
\label{sec:bridge_usage}
After training the Bridge, we attach the frozen module to the learnable Student to form an augmented Student that supports one-step distillation and one-step inference from the Teacher. This procedure follows the prior work summarized in \cref{section:one_step_distill} and is illustrated in \cref{fig:method}, in the top-right and bottom-right panels.

\section{Experiments}

\begin{table*}[!t]
\centering
\scriptsize
\renewcommand{\arraystretch}{1.15}
\arrayrulecolor{gray!70}
\caption{\textbf{Quantitative results.}
\textbf{A:} modern multi-step Teachers. \textbf{B:} compact one-step Students enabled by \emph{Cross-Space Distillation}.
Despite large gaps in resolution, VAE space, architecture, and generative mechanism, Bridge transfers knowledge from diverse modern Teachers into the same lightweight Student backbones while preserving one-step inference.
In \textbf{B}, the second column lists the Teacher used for each individual distillation run; initialized and merged rows summarize the corresponding Student family.
\textbf{Paras} are parameter counts in billions (B), and \textbf{NFE} denotes inference-time network evaluations.
\legendbox{initpink}{\textbf{Pink}} marks initialized Students, \legendbox{oursgreen}{\textbf{pale green}} marks individual Bridge-distilled Students, and \legendbox{mergegreen}{\textbf{stronger green}} marks checkpoints obtained by averaging all five distilled Students in the same family.
Across teacher families and metrics, Bridge \textbf{consistently improves} the initialized Students, and the merged checkpoints provide a simple post-training route to combine knowledge from multiple Teacher sources.
Higher values are better for all metrics.}
\vspace{-5pt}
\setlength{\tabcolsep}{2pt}
\renewcommand{\arraystretch}{1.18}

\begin{tabularx}{\textwidth}{@{}
L{0.13\textwidth} @{\hspace{6pt}} L{0.23\textwidth}
C{0.08\textwidth} C{0.055\textwidth}
*{5}{Y} @{}}
\toprule
\multicolumn{2}{@{}l}{\qquad\quad\textbf{Teacher}} &
\textbf{Paras} & \textbf{NFE} &
\multicolumn{5}{c}{\textbf{Metrics}} \\
\cmidrule(lr){5-9}
\textbf{Architect.} & \textbf{Model} &
\textbf{(B)} & \textbf{} &
\mHPSviii & \mHPSvii & \mIR & \mMPS & \mDPG \\
\midrule

% ---------------- Panel A ----------------
\multicolumn{9}{c}{\cellcolor{gray!10}\textbf{A. Multi-Step Teachers}} \\
\midrule

\multirow{2}{*}{Large U-Net} & SDXL~\cite{podell2024sdxl} &
2.57 & 50 &
9.25 & 28.36 & 0.66 & 13.72 & 74.00 \\
& Kolors~\cite{kolors} &
2.57 & 50 &
10.59 & 30.94 & 0.88 & 14.12 & 76.52 \\
\addlinespace[2pt]\midrule

DiT & PixArt-$\Sigma$-1024~\cite{chen2024pixartsigma} &
0.61 & 20 &
9.62 & 30.39 & 0.92 & 14.14 & 80.00 \\
\addlinespace[2pt]\midrule

\multirow{2}{*}{MM-DiT} & FLUX.2-klein-4B \cite{blackforest2025flux2} &
4.00 & 50 &
10.05 & 28.90 & 0.80 & 14.00 & 83.20 \\
& SD~3.5 Medium~\cite{sdv3} &
2.50 & 50 &
10.86 & 30.02 & 0.96 & 14.13 & 84.50 \\
\midrule

% ---------------- Panel B ----------------
\multicolumn{9}{c}{\cellcolor{gray!10}\textbf{B. Cross-Space Distillation}} \\
\midrule

\rowcolor{initpink}
\multicolumn{2}{@{}l}{\textbf{SD~1.5 Student}, DMD2~\cite{yin2024improved} (\textit{init.})} &
0.86 & 1 &
5.37 & 21.90 & -0.29 & 10.42 & 59.85 \\
\midrule

\rowcolor{oursgreen}
& SDXL~\cite{podell2024sdxl} &
0.86 & 1 &
9.04 & 27.37 & 0.30 & 11.93 & 63.00 \\
\rowcolor{oursgreen}
\multirow{-2}{*}{Large U-Net} & Kolors~\cite{kolors} &
0.86 & 1 &
9.33 & 27.38 & 0.34 & 11.66 & 64.46 \\
\addlinespace[2pt]\midrule

\rowcolor{oursgreen}
DiT & PixArt-$\Sigma$-1024~\cite{chen2024pixartsigma} &
0.86 & 1 &
8.65 & 26.43 & 0.30 & 11.94 & 63.43 \\
\addlinespace[2pt]\midrule

\rowcolor{oursgreen}
& FLUX.2-klein-4B~\cite{blackforest2025flux2} &
0.86 & 1 &
9.49 & 26.64 & 0.40 & 12.35 & 64.06 \\
\rowcolor{oursgreen}
\multirow{-2}{*}{MM-DiT} & SD~3.5 Medium~\cite{sdv3} &
0.86 & 1 &
9.42 & 28.30 & 0.62 & 12.96 & 65.75 \\
\midrule

\rowcolor{mergegreen}
\multicolumn{2}{@{}l}{\textbf{Merged (All 5 Teachers)}} &
0.86 & 1 &
\textbf{10.53} & \textbf{29.07} & \textbf{0.65} & 12.62 & \textbf{66.67} \\
\midrule

\rowcolor{initpink}
\multicolumn{2}{@{}l}{\textbf{SD~2.1 Student}, SiD \cite{zhou2024guided} \textit{(init.)}} &
0.86 & 1 &
6.42 & 23.74 & 0.10 & 11.29 & 61.33 \\
\midrule

\rowcolor{oursgreen}
& SDXL~\cite{podell2024sdxl} &
0.86 & 1 &
8.40 & 26.73 & 0.23 & 11.88 & 67.03 \\
\rowcolor{oursgreen}
\multirow{-2}{*}{Large U-Net} & Kolors~\cite{kolors} &
0.86 & 1 &
8.52 & 26.57 & 0.20 & 11.91 & 68.34 \\
\addlinespace[2pt]\midrule

\rowcolor{oursgreen}
DiT & PixArt-$\Sigma$-1024~\cite{chen2024pixartsigma} &
0.86 & 1 &
 8.42 & 28.29 & 0.44 & 12.20 & 68.51 \\
\addlinespace[2pt]\midrule

\rowcolor{oursgreen}
& FLUX.2-klein-4B~\cite{blackforest2025flux2} &
0.86 & 1 &
8.74 & 28.31 & 0.33 & 12.03 & 66.00 \\
\rowcolor{oursgreen}
\multirow{-2}{*}{MM-DiT} & SD~3.5 Medium~\cite{sdv3} &
0.86 & 1 &
8.64 & 29.11 & 0.45 & 12.92 & 68.52 \\
\midrule
\rowcolor{mergegreen}
\multicolumn{2}{@{}l}{\textbf{Merged (All 5 Teachers)}} &
0.86 & 1 &
\textbf{9.75} & \textbf{30.00} & \textbf{0.74} & 12.60 & 68.50 \\
\bottomrule
\end{tabularx}

\label{tab:main_results}
\vspace{-25pt}
\end{table*}

\label{sec:experiments}
\subsection{Experimental Setup}
\label{subsec:exp}
\myheading{Student Models ($\Student$).} We initialize the Student from one-step diffusion variants: SD~1.5 backbone from DMD2 \cite{yin2024improved,yin2024onestep} and SD~2.1 backbone from SiD-LSG \cite{zhou2024guided}. These models serve as representative legacy generators operating at $512 \times 512$ resolution.

\myheading{Teacher Models ($\Teacher$).} We evaluate our method under distillation settings using five state-of-the-art teachers operating at $1024 \times 1024$ resolution, including SDXL \cite{podell2024sdxl}, Kolors \cite{kolors}, PixArt-$\Sigma$ \cite{chen2024pixartsigma}, FLUX.2 [klein] 4B \cite{blackforest2025flux2}, and SD~3.5 Medium \cite{sdv3}. These models span different backbone architectures (U-Net, DiT, and MMDiT), and employ different latent configurations with varying VAE designs and channel dimensions (4, 16, and 32 channels).

\myheading{Bridge ($\Bridge_\phi$).}
Our Bridge module $\Bridge_\phi$ is a lightweight alignment network with approximately 5M trainable parameters.
As described in \cref{sec:bridge_arch}, $\Bridge_\phi$ is factorized into a frozen spatial prior and a learnable projection head.
The spatial prior is implemented by the first $n$ decoding blocks of the Student VAE decoder, denoted $\mathcal{D}_S^{(n)}$, which expands the Student latent grid to the Teacher spatial resolution.
In all experiments, we set $n=1$ and freeze $\mathcal{D}_S^{(1)}$.
The learnable projection head $g_\phi$ is implemented with SwinIR \cite{liang2021swinir} and maps the intermediate decoder features to a Teacher-compatible latent $\hat z_T$.
We train $\Bridge_\phi$ with the objectives in \cref{sec:bridge_objectives}. For Attention Fidelity, we use noising level $t=1$ and temperature $\tau=3.0$.
Ablations on architecture and objectives are provided in the supplementary material.

\myheading{Distillation Details.} We follow the distillation protocol of DMD2~\cite{yin2024improved}. All models are trained for 20 hours on 8$\times$ NVIDIA H100 GPUs with 80GB memory using the AdamW optimizer. Additional implementation details are provided in the supplementary material.

\myheading{Training Data.} To maximize \textit{training efficiency} and \textit{data accessibility}, we avoid relying on large-scale real-image datasets. Instead, we construct a static synthetic dataset of approximately 2M images by sampling text prompts from \textbf{JourneyDB} and \textbf{LAION} \cite{laion} and generating images with the corresponding teacher model. We use this synthetic dataset to train $\Bridge_\phi$ during space-bridge training. During distillation, we recycle the same dataset to compute an auxiliary adversarial objective, following the setup of DMD2.

\myheading{Metrics.} We evaluate Cross-Space Distillation using five widely adopted metrics: HPSv3 \cite{ma2025hpsv3}, HPSv2 \cite{wu2023hpsv2}, ImageReward \cite{xu2023imagereward}, MPS \cite{zhang2024mps}, and DPG Bench \cite{hu2024ella}. HPSv3 and HPSv2 assess human preference alignment on 12,000 and 3,200 prompts, respectively, while ImageReward is computed on 100 prompts. We further report MPS on the HPSv2 and ImageReward prompt sets to measure multi-dimensional scores. DPG Bench evaluates text-image alignment on 1,065 prompts. For all metrics, higher scores indicate better performance.

\subsection{Main Results}
\myheading{Cross-Architecture and Cross-Mechanism Distillation.} As demonstrated in \cref{tab:main_results}, our framework successfully facilitates distillation across architectures and diffusion frameworks, such as from DiT-based, flow matching models (\eg, SD~3.5 and FLUX.2) to UNet-based, noise-prediction models (\eg, SD~1.5). This versatility stems from two key design choices. First, our proposed Bridge explicitly neutralizes the Cross-Resolution and Cross-VAE constraints. Second, we adopt a mechanism-agnostic formulation by reparameterizing the student's prediction back to $x_0$ before sending it to the teacher, thereby eliminating discretization inconsistencies between the diffusion and flow matching formulations. Furthermore, we deliberately avoid layer-wise matching; supervision is imposed only at the prediction level, thereby eliminating structural constraints between the DiT and UNet backbones. Notably, even without architecture-based alignment, our method still achieves strong performance, demonstrating that once resolution and VAE mismatches are properly addressed, effective Cross-Space distillation becomes not only feasible but robust.

\myheading{Quantitative Results.}
\cref{tab:main_results} summarizes performance under five complementary metrics.
Across all teacher models, Cross-Space Distillation improves the one-step Students on these metrics, suggesting that the gains are not confined to a single evaluator.
For instance, using SD~3.5 Medium as the Teacher increases the SD~1.5 Student from 5.37 to 9.42 on HPSv3 and from $-0.29$ to 0.62 on ImageReward, with matching improvements on HPSv2, MPS, and DPG Bench. Overall, these results suggest higher perceived quality and stronger prompt adherence while preserving the same one-step inference budget.

\begin{figure}[!t]
    \centering
    \includegraphics[width=\linewidth]{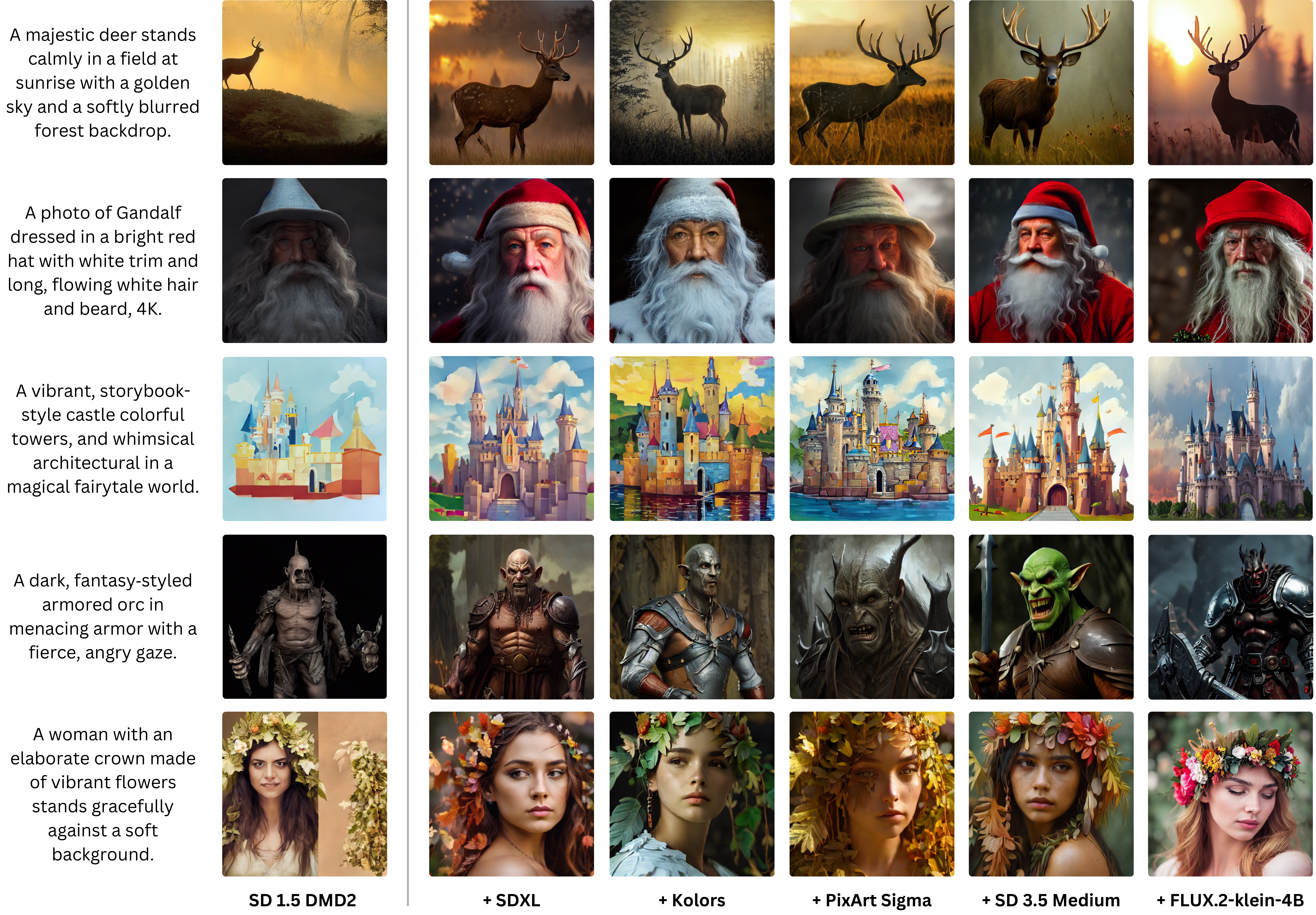}
    \vspace{-7pt}
    \caption{\textbf{Qualitative Comparison of Cross-Space Distillation.} We show samples from the initialized one-step student SD~1.5 DMD2 at 512$\times$512 and from one-step students distilled with our method using different 1024$\times$1024 teachers. Columns labeled \textbf{+Teacher} denote the student distilled from that teacher.}
    \label{fig:qual}
\end{figure}

\myheading{Qualitative Results.}
We report qualitative examples in \cref{fig:qual}. Across a diverse set of prompts, our approach produces images with clearer textures, stronger structural consistency, and more faithful local details than the initialized baseline. In contrast, the baseline more frequently exhibits softness, broken fine structure, and occasional anatomical artifacts. These examples visually support the quantitative gains by showing that Bridge enabled distillation improves perceptual sharpness and coherence while preserving one-step generation.

\subsection{Post-Training Improvement via Model Merging}
As an optional post-training refinement, we investigate \textbf{model merging} across students distilled from different teachers. This experiment is \emph{not} part of the core Bridge design; rather, it serves as a practical extension that can further improve the final student \textbf{without changing the model architecture, parameter count, or inference cost}. Since all distilled checkpoints within each student family share the same backbone and parameterization, they can be combined through \textbf{direct parameter averaging}.

Given a set of distilled one-step student checkpoints
$\mathcal{S}=\{\theta_1,\theta_2,\ldots,\theta_{|\mathcal{S}|}\}$,
we form a merged checkpoint by taking the element-wise mean of the parameters:
\begin{equation}
    \theta_{\text{merged}}
    =
    \frac{1}{|\mathcal{S}|}
    \sum_{i=1}^{|\mathcal{S}|}\theta_i.
\end{equation}
In our experiments, we merge the five students obtained by distilling from the five different teacher models in the main paper. This simple procedure consolidates multiple teacher-specific distilled checkpoints into a \textbf{single student checkpoint} while preserving the same inference-time NFE and model size.

\cref{tab:main_results} reports the merged results as highlighted rows under both SD~1.5 and SD~2.1 student families. Parameter averaging yields a stronger \emph{overall} metric profile than the initialized student and is often competitive with, or better than, the individual distilled checkpoints on HPSv3, HPSv2, and ImageReward. For example, the merged SD~1.5 checkpoint improves \textbf{\textit{HPSv3 from 5.37 to 10.53}} and \textbf{\textit{HPSv2 from 21.90 to 29.07}}, while the merged SD~2.1 checkpoint improves \textbf{\textit{HPSv3 from 6.42 to 9.75}} and \textbf{\textit{HPSv2 from 23.74 to 30.00}}. These results suggest that model merging can serve as a lightweight post-training enhancement for Cross-Space Distillation.

This model merging presents an \textit{unique benefit} of our technique. Normally, it is hard to combine the image-generation capabilities of state-of-the-art models like Flux, SD~3.5, or Kolors, given their different network architectures and latent spaces. However, our technique distills them into student models with the same backbone and latent space, and combining knowledge across these student models is trivial. It provides a pathway to \textbf{unify generation capabilities from multiple model sources}. With more teacher models and/or stronger distillation techniques beyond DMD2, we expect an even stronger merged student model that can outperform every single teacher model.

\begin{table}[!t]
\centering
\scriptsize
\setlength{\tabcolsep}{3pt}
\renewcommand{\arraystretch}{1.15}
\arrayrulecolor{gray!70}

\caption{\textbf{Pruning versus Bridge enabled distillation.} We prune SD~3.5 Medium to 30\% sparsity with OBS-Diff \cite{zhu2025obsdiff} and use it to initialize a one-step Student, then distill from the full SD~3.5 Medium Teacher using the same setup as \cref{subsec:exp}. Bridge enabled distillation yields higher preference scores and cleaner details, while pruning based initialization more often produces blur and artifacts.}

\begin{tabular}{@{}lcccccc@{}}
\toprule
\textbf{Setting} & \textbf{Params (B)} & \textbf{HPSv3} & \textbf{HPSv2} & \textbf{IR} & \textbf{MPS} & \textbf{DPG} \\
\midrule
Prune (30\%) & 1.5  & 2.76 & 21.94 & -0.36 & 9.54  & 55.24 \\
Ours         & 0.86 & 9.42 & 28.30 & 0.62  & 12.96 & 65.75 \\
\bottomrule
\end{tabular}
\label{tab:prune}
\vspace{-10pt}
\end{table}

\subsection{Additional Analyses}\label{sec:additional_analyses}

\myheading{Student Initialization.}
We compare two Student initializations: one-step SD 1.5 DMD2 checkpoint and standard multi-step SD 1.5 Teacher weights. With SD~3.5 Medium as the Teacher (\cref{tab:init}), DMD2-init is slightly better in HPSv2, while SD 1.5 is slightly better on other metrics. This suggests initialization has minimal influence on performance in our framework. To maximize training speed and efficiency, we opt to initialize DMD2 as students.

\begin{figure}[!t]
    \centering
    \vspace{-2pt}
    \includegraphics[width=\linewidth]{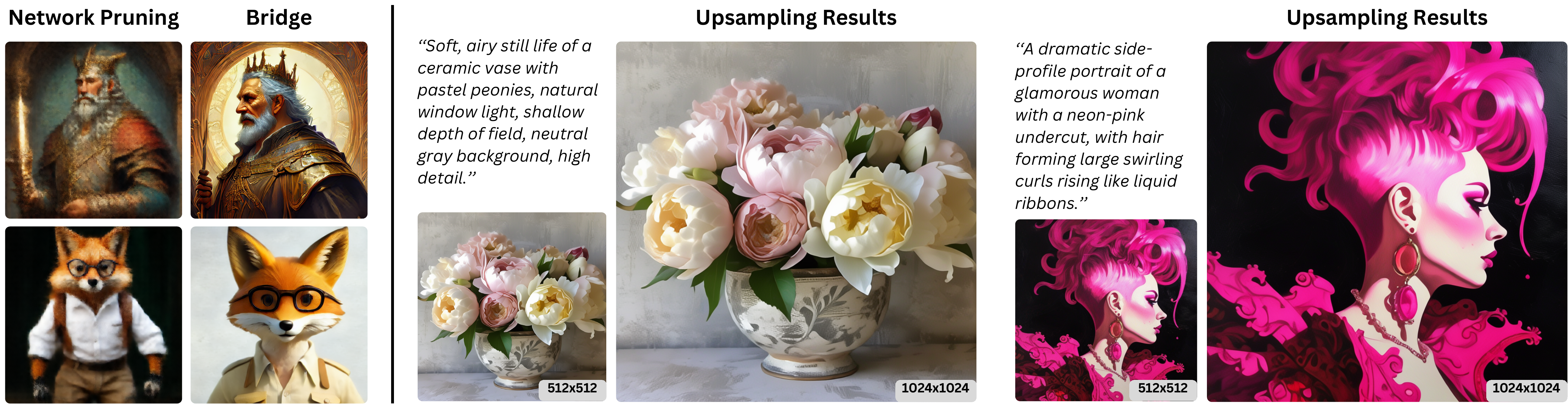}
    \vspace{-2pt}
    \caption{\textbf{Qualitative comparisons.}
    \textbf{(Left) Pruning vs.\ Bridge distillation.}
    We compare a pruned baseline (1.5B) against our Bridge-enabled compact student (0.6B), generated with identical prompts and NFE $=1$. \textbf{(Right) Resolution upgrade at inference time.} A 512$\times$512 sample from the same one-step student is mapped by \textbf{Bridge} into the teacher latent space and decoded with the teacher decoder to produce a 1024$\times$1024 output.}
    \label{fig:qual_combo}
\end{figure}

\myheading{Comparison with Network Pruning.}
To benchmark against parameter reduction, we compare the Bridge with network pruning. Using OBS-Diff~\cite{zhu2025obsdiff}, we prune SD~3.5 Medium to 30\% sparsity (2.5B $\rightarrow$ 1.5B) and use it to initialize the one-step Student $\Student$, then distill from the full SD~3.5 Medium Teacher following \cref{sec:experiments}. As shown in \cref{tab:prune} and \cref{fig:qual_combo}, with the same NFE=1 budget, our approach yields notably sharper outputs and stronger preferences, while pruned-model distillation tends to introduce blur and artifacts.

\myheading{Resolution Upgrade with Bridge.}
Once trained, our Bridge $\Bridge_\phi$ can also serve as a plug-and-play inference module that upgrades a low-resolution generator to higher-resolution outputs without retraining the generator itself. As shown in the right pane of \cref{fig:qual_combo}, the Bridge synthesizes semantically consistent high-frequency details from low-resolution latents. This demonstrates that $\Bridge_\phi$ generalizes to unseen inputs and provides a robust, cross-resolution mapping.

\begin{table}[t]
\centering
\scriptsize
\setlength{\tabcolsep}{3pt}
\renewcommand{\arraystretch}{1.15}
\arrayrulecolor{gray!70}
\caption{\textbf{Student initialization.}
We compare initializing the Student from a one-step SD 1.5 DMD2 checkpoint versus standard multi-step SD 1.5 weights, with both distilled from the same SD~3.5 Medium Teacher.
Results are similar across metrics, so we use DMD2 initialization for subsequent experiments due to better training efficiency.}
\vspace{-5pt}
\renewcommand{\arraystretch}{1.15}
\arrayrulecolor{gray!70}
\begin{tabular}{@{}lcccccc@{}}
\toprule
\textbf{Setting} & \textbf{Iteration} & \textbf{HPSv3} & \textbf{HPSv2} & \textbf{IR} & \textbf{MPS} & \textbf{DPG}  \\
\midrule
SD 1.5 Init. & 9k & 9.22 & \textbf{28.44} & \textbf{0.66} & \textbf{13.08} & \textbf{66.89} \\
SD 1.5 DMD2 Init. & 3k & \textbf{9.42} & 28.30 & 0.62  & 12.96 & 65.75 \\
\bottomrule
\end{tabular}
\label{tab:init}
\end{table}

\section{Related Work}

\myheading{Diffusion and Flow.}
\textit{Diffusion models} have become the dominant paradigm for high-quality image generation. Early works such as DDPM \cite{ddpm} and score-based generative models \cite{score_sde} introduced a stochastic forward corruption process and a learned reverse denoising process for generative modeling. Subsequent improvements have significantly accelerated sampling through more efficient solvers \cite{ddim,dpm_solver,edm} and improved training formulations \cite{progressive_distillation}.

Recently, \textit{flow matching models} \cite{flow_matching,rectified_flow,cfm_ot} provide an alternative continuous-time formulation that directly learns the velocity field transporting samples from noise to data. Modern text-to-image systems such as SD~3.5 \cite{sdv3} and Flux \cite{blackforest2024flux1,blackforest2025flux2} build on these advances with large-scale backbones and high-resolution latent representations, achieving strong visual fidelity and prompt alignment. However, these models typically require multiple network evaluations during inference, resulting in substantial computational overhead.

\myheading{Distillation for Fast Diffusion Sampling.}
To reduce inference cost, a large body of work distills multi-step diffusion models into fast generators with significantly fewer sampling steps. Distribution-based distillation methods train a Student generator to match the sample distribution implied by a pretrained Teacher model. Representative approaches include variational score distillation and related methods \cite{vsd,sb1,f_distill,snoopi,yin2024improved,yin2024onestep}, which use score differences between Teacher and Student predictions as training signals. Other approaches explore consistency-style distillation \cite{lcm,tcm} or trajectory-based methods that directly approximate multi-step sampling dynamics.

Several works further combine distribution matching with adversarial supervision to improve perceptual quality and reduce the gap between fast generators and their multi-step Teachers \cite{add,ladd}. While these approaches substantially reduce the number of inference steps, they typically assume that the Teacher and Student operate within the same latent representation and spatial resolution. Consequently, the distilled model often inherits the architectural scale or latent structure of the Teacher, limiting its suitability for lightweight deployment.

In this paper, we study \textbf{Cross-Space Distillation}, introducing a lightweight alignment module that maps Student representations into the Teacher latent space, enabling standard distillation objectives without modifying the Student architecture. This allows compact backbones such as Stable Diffusion 1.5 \cite{sd1} to inherit the generative capability of modern high-capacity models while maintaining ecosystem compatibility.

\section{Conclusion}
\label{sec:conclusion}

We identify a common but restrictive assumption in distribution-based one-step distillation: Teacher and Student are expected to share the same latent representation.
We relax this assumption through \textbf{Cross-Space Distillation} and introduce \textbf{Bridge} ($\Bridge_\phi$), a lightweight latent interface that maps Student latents into the Teacher space, making standard one-step distillation objectives applicable under latent-resolution and VAE mismatch.
Across modern Teachers and compact Students, Bridge yields strong preference gains while preserving one-step inference, the original Student backbone, and compatibility with widely used deployment ecosystems. Beyond training, Bridge can also be reused at inference time to map low-resolution Student samples onto the Teacher latent grid for higher-resolution decoding. We hope this work motivates broader exploration of distillation and model reuse across heterogeneous latent representations, and inspires future research on alignment interfaces that unlock knowledge transfer across mismatched latent spaces.

\newpage
\bibliographystyle{splncs04}
\bibliography{main}

% Optional supplementary material should be submitted separately as ESM in a
% `Supplementary_Material` directory, not appended to the proceedings PDF.
\newpage
\newcommand{\SUPPLEMENTARYINPUT}{}
% ECCV 2026 supplementary template
% Standalone supplementary file matching the ECCV 2026 submission format.
% Submission ID is kept consistent with the main paper.

\ifdefined\SUPPLEMENTARYINPUT
\else
\ifdefined\pdfminorversion
  \pdfminorversion=7
\fi
\documentclass[runningheads]{llncs}
\PassOptionsToPackage{table}{xcolor}

% ---------------------------------------------------------------
% ECCV 2026 package
% Keep review mode and submission ID for the submission version.
%\usepackage[review,year=2026,ID=7574]{eccv}
% For camera-ready, switch to:
\usepackage{eccv}

% Optional mobile-friendly layout
% \usepackage[mobile]{eccv}

% ---------------------------------------------------------------
% Common ECCV abbreviations
\usepackage{eccvabbrv}

% ---------------------------------------------------------------
% Packages copied from the main file for compatibility
\usepackage{array}
\usepackage{graphicx}
\usepackage{booktabs}
\usepackage{makecell}
\usepackage{amsmath, amssymb, amsfonts}
\usepackage{bm}
\usepackage{tabularx}
\usepackage{colortbl}
\usepackage{multirow}
\usepackage[table]{xcolor}
\usepackage{pifont}
\usepackage{xkcdcolors}
\usepackage[most]{tcolorbox}
\usepackage{orcidlink}
\usepackage{enumitem}
\usepackage{hyperref}

% ---------------------------------------------------------------
% Column types
\newcolumntype{L}[1]{>{\raggedright\arraybackslash}p{#1}}
\newcolumntype{Y}{>{\centering\arraybackslash}X}
\newcolumntype{C}[1]{>{\centering\arraybackslash}p{#1}}

% ---------------------------------------------------------------
% Custom colors from the main file
\definecolor{burntorange}{RGB}{204, 85, 0}
\definecolor{crimson}{RGB}{220, 20, 60}
\definecolor{teal}{RGB}{0, 128, 128}
\definecolor{royalblue}{RGB}{65, 105, 225}
\definecolor{magenta}{RGB}{255, 0, 255}
\definecolor{forestgreen}{RGB}{34, 139, 34}
\definecolor{deeppink}{RGB}{255, 20, 147}
\definecolor{chocolate}{RGB}{210, 105, 30}
\definecolor{darkviolet}{RGB}{148, 0, 211}
\definecolor{cerulean}{RGB}{0, 123, 167}
\definecolor{olive}{RGB}{128, 128, 0}

\definecolor{baselinebg}{HTML}{F2F2F2}
\definecolor{headerbg}{HTML}{D9EAD3}
\definecolor{fluxbg}{HTML}{E2EFDA}

% ---------------------------------------------------------------
% Marks
\newcommand{\cmark}{\textcolor{green!70!black}{\ding{51}}}
\newcommand{\xmark}{\textcolor{red!80!black}{\ding{55}}}

% ---------------------------------------------------------------
% Comment macros from the main file
\newcommand{\anh}[1]{\textcolor{burntorange}{(Anh Tran: #1)}}
\newcommand{\trung}[1]{\textcolor{darkviolet}{(Trung: #1)}}
\newcommand{\duc}[1]{\textcolor{chocolate}{(Duc: #1)}}
\newcommand{\khoi}[1]{\textcolor{magenta}{(Khoi: #1)}}
\newcommand{\ngan}[1]{\textcolor{cerulean}{(Ngan: #1)}}
\newcommand{\kien}[1]{\textcolor{olive}{(Kien: #1)}}
\newcommand{\quan}[1]{\textcolor{crimson}{(Quan Dao: #1)}}
\newcommand{\viet}[1]{\textcolor{teal}{(Viet: #1)}}
\newcommand{\phong}[1]{\textcolor{royalblue}{(Phong: #1)}}
\newcommand{\cuong}[1]{\textcolor{forestgreen}{(Cuong: #1)}}

% ---------------------------------------------------------------
% Notation consistency
\newcommand{\Teacher}{\mathcal{T}}
\newcommand{\Student}{\mathcal{S}}
\newcommand{\Bridge}{\mathcal{B}}
\newcommand{\Loss}{\mathcal{L}}
\newcommand{\LatentT}{\mathcal{Z}_\Teacher}
\newcommand{\LatentS}{\mathcal{Z}_\Student}
\newcommand{\spaceT}{\mathcal{M}_\Teacher}
\newcommand{\spaceS}{\mathcal{M}_\Student}

\newcommand{\mHPSviii}{\textbf{\mbox{HPSv3}}}
\newcommand{\mHPSvii}{\textbf{\mbox{HPSv2}}}
\newcommand{\mIR}{\textbf{\mbox{IR}}}
\newcommand{\mMPS}{\textbf{\mbox{MPS}}}
\newcommand{\mDPG}{\textbf{\mbox{DPG}}}

\newcommand{\myheading}[1]{\vspace{2mm}\noindent{\textbf{#1}}}

\newcommand{\finding}[2]{
    \begin{tcolorbox}[
        colback=white!90!gray,
        colframe=teal!60!black,
        arc=5pt,
        boxsep=5pt,
        left=10pt,
        right=10pt,
        top=2pt,
        bottom=2pt,
        boxrule=0.8pt,
        drop shadow=gray!50!white,
        enhanced jigsaw
    ]
    \vspace{-0.1cm}
        \paragraph{\textbf{\textit{}}} #2
    \end{tcolorbox}
    \vspace{-0.1cm}
}
\fi

\renewcommand{\thesection}{\Alph{section}}
\renewcommand{\thefigure}{A\arabic{figure}}
\renewcommand{\thetable}{A\arabic{table}}
\renewcommand{\theequation}{A\arabic{equation}}

\ifdefined\SUPPLEMENTARYINPUT
\clearpage
\setcounter{section}{0}
\setcounter{figure}{0}
\setcounter{table}{0}
\setcounter{equation}{0}
\else
\begin{document}
\fi

\thispagestyle{empty}
\vspace*{6mm}
\begin{center}
{\large\bfseries
Cross-Space Distillation:\\
Teaching One-Step Students with Modern Diffusion Teachers\par}
\vspace{2mm}
{\large\bfseries -- Supplementary Materials --\par}
\end{center}
\vspace{10mm}
\ifdefined\SUPPLEMENTARYINPUT
\else
\setcounter{page}{1}
\fi

% Uncomment if you prefer appendix-style lettering for sections.
% \appendix
\myheading{Overview.} This supplementary material expands the main paper with additional implementation details, Bridge design ablations, efficiency analyses, and extended qualitative results. Beyond supporting reproducibility, these results further illustrate the practical strengths of Cross-Space Distillation: Bridge is lightweight, reusable across diverse teachers, efficient to train as a post-training module, and effective in transferring strong visual priors into compact one-step students. We first report implementation details for Bridge training and downstream distillation, then provide additional details on Bridge design, efficiency, usage, and ablations, and finally include extended qualitative comparisons. We conclude by discussing the current scope of the method and promising directions beyond the main paper.

\section{Implementation and Hyperparameter Details}

We summarize the implementation details for both \emph{Bridge training} and \emph{downstream one-step distillation} in \cref{tab:training_settings}. Across all teacher models, we use the same Bridge architecture and modify only the final projection layer to match the latent dimensionality of the target teacher. The Bridge is trained from scratch, without pretrained initialization. In our setup, both stages fit on a single 8$\times$H100 (80GB) node, making the overall pipeline practical as a post-training recipe rather than a full model pretraining procedure. Bridge training optimizes only the lightweight alignment module, whereas the subsequent student distillation follows a standard one-step training setup with minor batch-size adjustments across teachers.

\begin{table*}[t]
\centering
\scriptsize
\setlength{\tabcolsep}{3pt}
\renewcommand{\arraystretch}{1.12}
\caption{\small Implementation details, hyperparameters, and compute settings for Bridge training and downstream one-step distillation of students from different teachers. Across all teachers, Bridge uses the same architecture and changes only its output dimensionality to match the target teacher latent space. Both stages fit on a single 8$\times$H100 (80GB) node, making the overall pipeline practical as a reusable post-training recipe rather than a full model pretraining procedure. Reported training time excludes text-encoding overhead. We use FLUX.2 and SD~3.5 as abbreviations for FLUX.2-Klein-4B and SD~3.5 Medium, respectively.}
\label{tab:training_settings}

\begin{tabularx}{\textwidth}{@{}L{0.35\textwidth}*{5}{>{\centering\arraybackslash}X}@{}}
\toprule
\multicolumn{6}{c}{\textbf{Bridge Settings}} \\
\midrule
Architecture                     & \multicolumn{5}{c}{SwinIR} \\
Number of synthesized images     & \multicolumn{5}{c}{2M} \\
Trainable parameters             & \multicolumn{5}{c}{5M} \\
Loss weight $\alpha$             & \multicolumn{5}{c}{1.0} \\
Loss weight $\beta$              & \multicolumn{5}{c}{1.0} \\
Learning rate for $\Bridge$      & \multicolumn{5}{c}{1e-4} \\
Optimizer                        & \multicolumn{5}{c}{Adam ($\beta_1=0$, $\beta_2=0.999$, $\epsilon=10^{-8}$)} \\
Compute                          & \multicolumn{5}{c}{8$\times$H100 (80GB)} \\
Batch size per GPU               & \multicolumn{5}{c}{32} \\
Training time (hours)            & \multicolumn{5}{c}{8} \\
\midrule
\multicolumn{6}{c}{\textbf{Distillation Settings}} \\
\midrule
Teacher model
& \makecell[c]{SDXL}
& \makecell[c]{Kolors}
& \makecell[c]{PixArt-$\sigma$}
& \makecell[c]{FLUX.2}
& \makecell[c]{SD~3.5} \\
\midrule
Generative paradigm
& Diffusion
& Diffusion
& Diffusion
& \makecell[c]{Flow}
& \makecell[c]{Flow} \\
Teacher latent channels
& 4 & 4 & 4 & 32 & 16 \\
Student learning rate
& \multicolumn{5}{c}{1e-6} \\
\makecell[l]{Auxiliary score model\\learning rate}
& \multicolumn{5}{c}{5e-4} \\
Discriminator learning rate
& \multicolumn{5}{c}{5e-7} \\
Adversarial loss weight $w_{GAN}$
& \multicolumn{5}{c}{0.5} \\
\makecell[l]{Distribution-matching loss\\weight $w_{VSD}$}
& \multicolumn{5}{c}{1.0} \\
Optimizer
& \multicolumn{5}{c}{Adam ($\beta_1=0$, $\beta_2=0.999$, $\epsilon=10^{-8}$)} \\
Compute
& \multicolumn{5}{c}{8$\times$H100 (80GB)} \\
Batch size per GPU
& 32 & 32 & 32 & 16 & 32 \\
Training time (hours)
& 20 & 20 & 20 & 20 & 20 \\
\bottomrule
\end{tabularx}
\end{table*}

\section{Bridge Design, Efficiency, and Usage Details}

\subsection{Architectural Design}
The ablations in \cref{fig:bridge_ablation} compare Bridge architectures under the same training budget, with 5M trainable parameters and 10K training iterations. The results confirm the intended ordering of the architectural variants: MLP performs worst, UNet provides a stronger baseline, SwinIR improves further over UNet, and SwinIR with the proposed Spatial Prior achieves the best overall reconstruction fidelity. These results support our design choice: a stronger image-restoration-style backbone is more effective for latent-space alignment, and the frozen Spatial Prior provides an additional gain beyond backbone choice alone.

\subsection{Training Objectives}
\begin{figure}[!t]
    \centering
    \includegraphics[width=\linewidth]{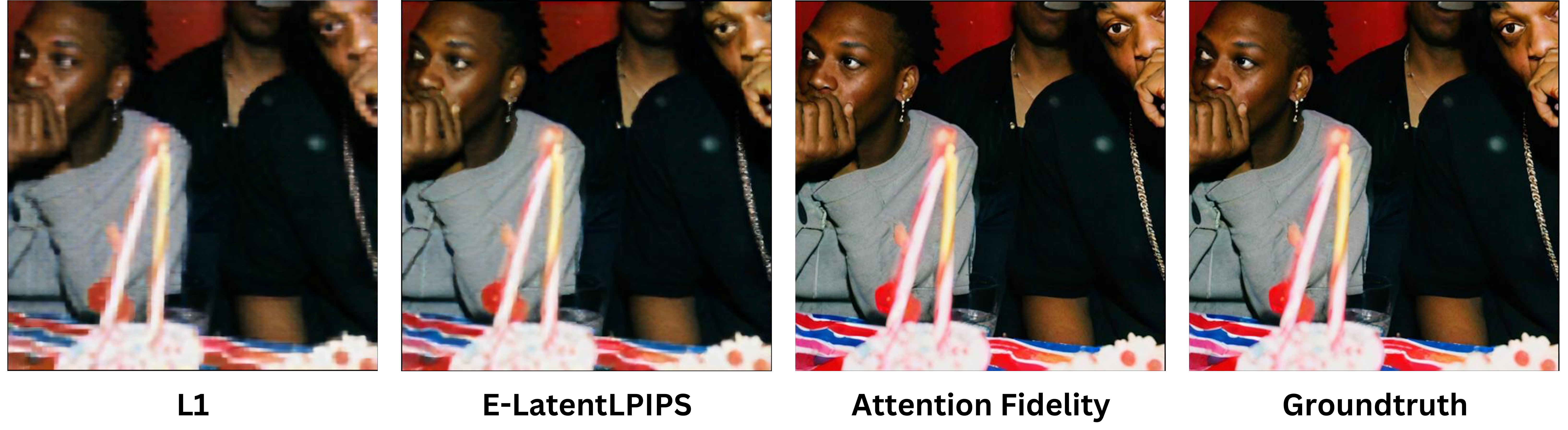}
    \caption{Qualitative comparison of training objectives for $\Bridge$. From left to right: reconstructions trained with $\ell_1$ only, $\ell_1$ + E-LatentLPIPS, and $\ell_1$ + Attention Fidelity, followed by the ground-truth reconstruction. All variants use the SwinIR backbone and are trained to map SD~2.1 latents to the SDXL latent space. While $\ell_1$ preserves the overall structure, it produces noticeably smooth reconstructions, and E-LatentLPIPS provides only marginal visual improvement. The proposed Attention Fidelity loss better preserves local structure and fine facial details, yielding reconstructions that are visually closest to the ground truth. Please zoom in for closer inspection.}
    \label{fig:loss}
\end{figure}

\cref{fig:loss} provides a qualitative comparison of the training objectives used for learning $\Bridge$. Using only the $\ell_1$ reconstruction loss preserves the coarse object layout, but the reconstructed image remains visibly over-smoothed, with weakened local contrast and softened facial structure. Adding E-LatentLPIPS yields only limited perceptual improvement over $\ell_1$ alone. In contrast, incorporating the proposed \emph{Attention Fidelity} loss produces substantially sharper and more faithful reconstructions, especially in fine-grained regions such as the eyes, nose contour, and surrounding fur. These visual differences are consistent with the quantitative ablation in the main paper, where Attention Fidelity improves reconstruction quality over both $\ell_1$ and E-LatentLPIPS.

\subsection{Efficiency}
\cref{tab:bridge_overhead} reports the average runtime and peak memory overhead introduced by Bridge during inference. All measurements are averaged over 500 iterations on a single H100 80GB GPU with batch size $1$. We report results for the three Bridge variants that map the SD~1.5 latent space to the latent spaces of SDXL, SD~3.5 Medium, and FLUX.2-klein-4B; Kolors and PixArt-$\sigma$ share the same VAE as SDXL and therefore use the same Bridge configuration.

Overall, Bridge introduces only a modest increase in runtime and memory over the base SD~1.5 pipeline. The overhead grows with the target latent dimensionality, as expected, but remains manageable even for the 32-channel FLUX latent space. These measurements support the practical use of Bridge as a lightweight latent-space interface rather than a heavy second-stage generator.

\begin{table}[t]
\scriptsize
\centering
\caption{\small Runtime and peak memory overhead introduced by Bridge over the base SD~1.5 pipeline. Measurements are averaged over 500 iterations on a single H100 80GB GPU with batch size 1. The results show that Bridge adds only modest inference overhead while enabling alignment to higher-dimensional teacher latent spaces.}
\label{tab:bridge_overhead}
\setlength{\tabcolsep}{5pt}
\renewcommand{\arraystretch}{1.08}
\begin{tabular}{lccc}
\toprule
Setting & Latent ch. & Runtime (ms) & Peak GPU Memory (GB) \\
\midrule
Base SD~1.5 (No $\mathcal{B}$)   & 4  & 26.48          & 3.96 \\
+$\mathcal{B}$ to SDXL            & 4  & 32.49 (+6.01)  & 4.50 (+0.54) \\
+$\mathcal{B}$ to SD~3.5 Medium   & 16 & 33.93 (+7.45)  & 4.51 (+0.55) \\
+$\mathcal{B}$ to FLUX.2-klein-4B & 32 & 44.49 (+18.01) & 4.74 (+0.78) \\
\bottomrule
\end{tabular}
\end{table}

\subsection{Usage During Distillation}
Unless otherwise stated, $\Bridge$ is kept frozen throughout downstream Cross-Space distillation. We additionally explored jointly updating $\Bridge$ together with the Student during distillation, but observed unstable optimization and, in some cases, training collapse. We therefore freeze $\Bridge$ in all downstream distillation experiments. This design stabilizes training, reduces memory usage, and preserves $\Bridge$ as a pretrained latent-space interface rather than a second trainable generator component.

\subsection{Inference-Time Resolution Upgrade}
\begin{figure}[!t]
    \centering
    \includegraphics[width=\linewidth]{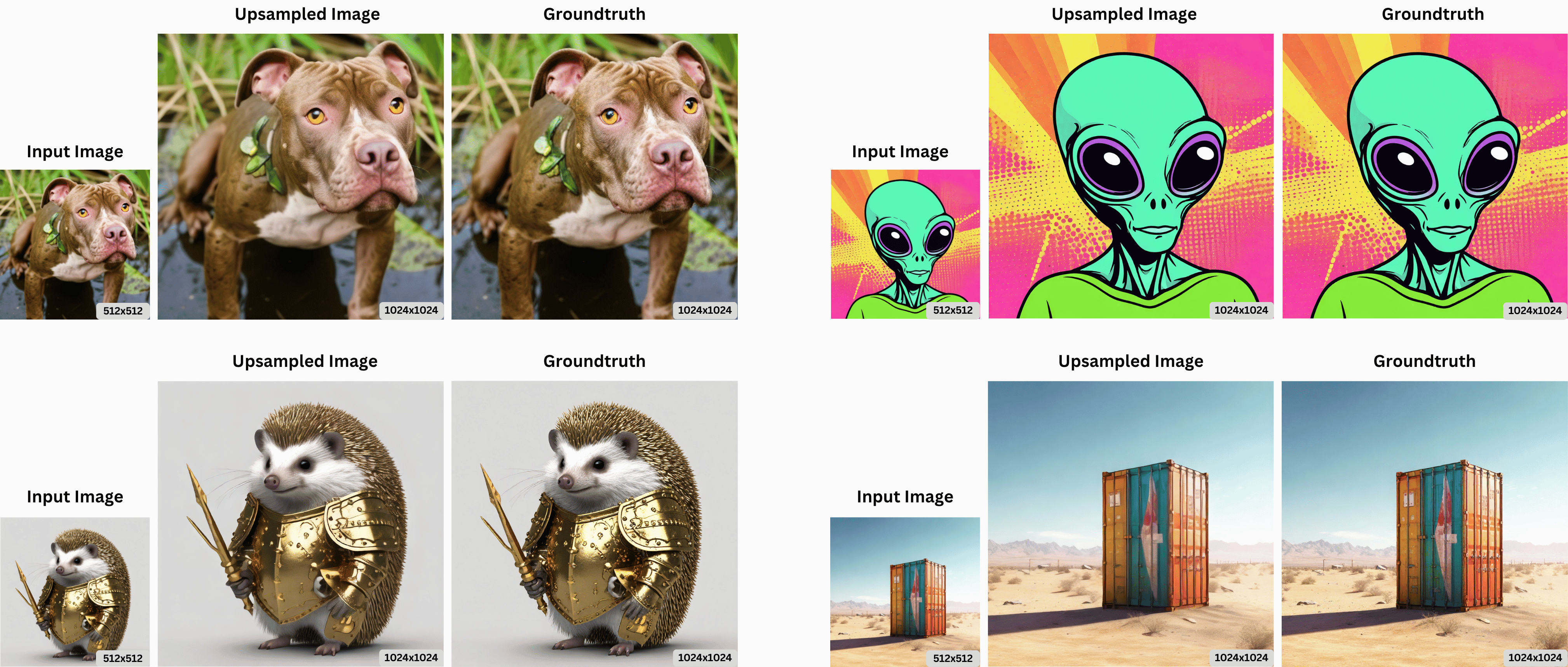}
    \caption{Qualitative examples of inference-time resolution upgrade with Bridge for the SD~1.5 Student distilled from the SD~3.5 Medium Teacher. Given low-resolution inputs at $512\times512$, Bridge predicts teacher-compatible high-resolution latents at $1024\times1024$ that closely match the target reconstructions, while preserving fine structures, textures, and global consistency across diverse scenes. This experiment evaluates Bridge in the same $512 \rightarrow 1024$ setting studied throughout the paper.}    \label{fig:recons}
\end{figure}

We provide additional qualitative examples of inference-time resolution upgrade using Bridge in \cref{fig:recons}. Starting from low-resolution inputs, Bridge predicts teacher-compatible high-resolution latents that can be decoded at $1024\times1024$ while preserving structure, texture, and global coherence. This shows that Bridge can be reused at inference time as a lightweight latent-space interface, beyond its role in downstream distillation.

This inference setting should be interpreted in the same scope as the main paper: Bridge maps a low-resolution Student latent onto the Teacher latent grid under Cross-Resolution and Cross-VAE mismatch. Our focus is therefore the practical $512 \rightarrow 1024$ setting studied throughout the paper, which matches the Student--Teacher pairs used in Cross-Space Distillation. Larger output scales and arbitrary output resolutions are beyond the scope of this work.

\subsection{Bridge Ablation Studies}
\label{sec:supp_ablation}
\begin{figure}[t]
\begin{minipage}[t]{0.48\linewidth}
\vspace{0pt}
\centering
\scriptsize
\renewcommand{\arraystretch}{1.15}
\arrayrulecolor{gray!70}
\begin{tabularx}{\linewidth}{Xccc}
\toprule[0.6pt]
Framework & L1 $\downarrow$ & PSNR $\uparrow$ & SSIM $\uparrow$ \\
\midrule[0.3pt]
Baseline & 0.048 & 22.6 & 0.68 \\
~+ Attention Fidelity & 0.037 & 24.39 & 0.73  \\
\rowcolor{green!10}
~+ Spatial Prior & \textbf{0.035} & \textbf{24.97} & \textbf{0.75} \\
\bottomrule[0.6pt]
\end{tabularx}
\vspace{2mm}
(a) Components of proposed framework
\vspace{2mm}
\begin{tabularx}{\linewidth}{Xccc}
\toprule[0.6pt]
Objectives & L1 $\downarrow$ & PSNR $\uparrow$ & SSIM $\uparrow$ \\
\midrule[0.3pt]
L1 & 0.041 & 23.58 & 0.70 \\
E-LatentLPIPS & 0.042 & 23.62 & 0.71 \\
\rowcolor{green!10}
Attention Fidelity & \textbf{0.035} & \textbf{24.97} & \textbf{0.75} \\
\bottomrule[0.6pt]
\end{tabularx}
\vspace{2mm}
(b) Objectives
\vspace{2mm}
\begin{tabularx}{\linewidth}{Xccc}
\toprule[0.6pt]
Architectures & L1 $\downarrow$ & PSNR $\uparrow$ & SSIM $\uparrow$ \\
\midrule[0.3pt]
MLP & 0.075 & 19.46 & 0.60 \\
UNet & 0.040 & 23.61 & 0.68 \\
SwinIR~\cite{liang2021swinir} & 0.037 & 24.39 & 0.73 \\
\rowcolor{green!10}
~ + Spatial Prior & \textbf{0.035} & \textbf{24.97} & \textbf{0.75} \\
\bottomrule[0.6pt]
\end{tabularx}
\vspace{2mm}
(c) Architectures
\arrayrulecolor{black}
\end{minipage}
\hfill
\begin{minipage}[t]{0.48\linewidth}
\vspace{0pt}
\scriptsize
\centering
\includegraphics[width=.8\linewidth]{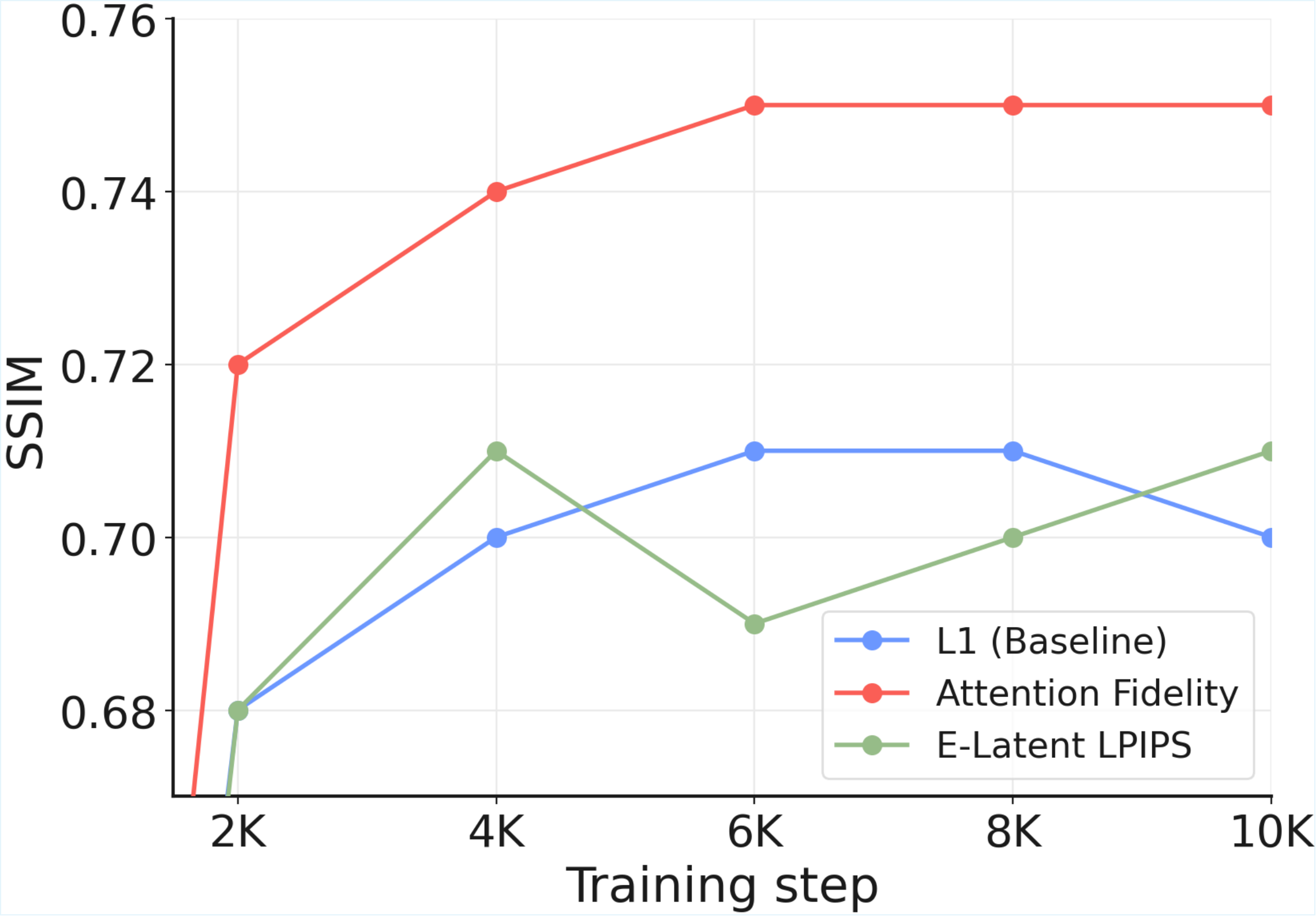}
{\scriptsize (d) Convergence across objectives.}
\includegraphics[width=.8\linewidth]{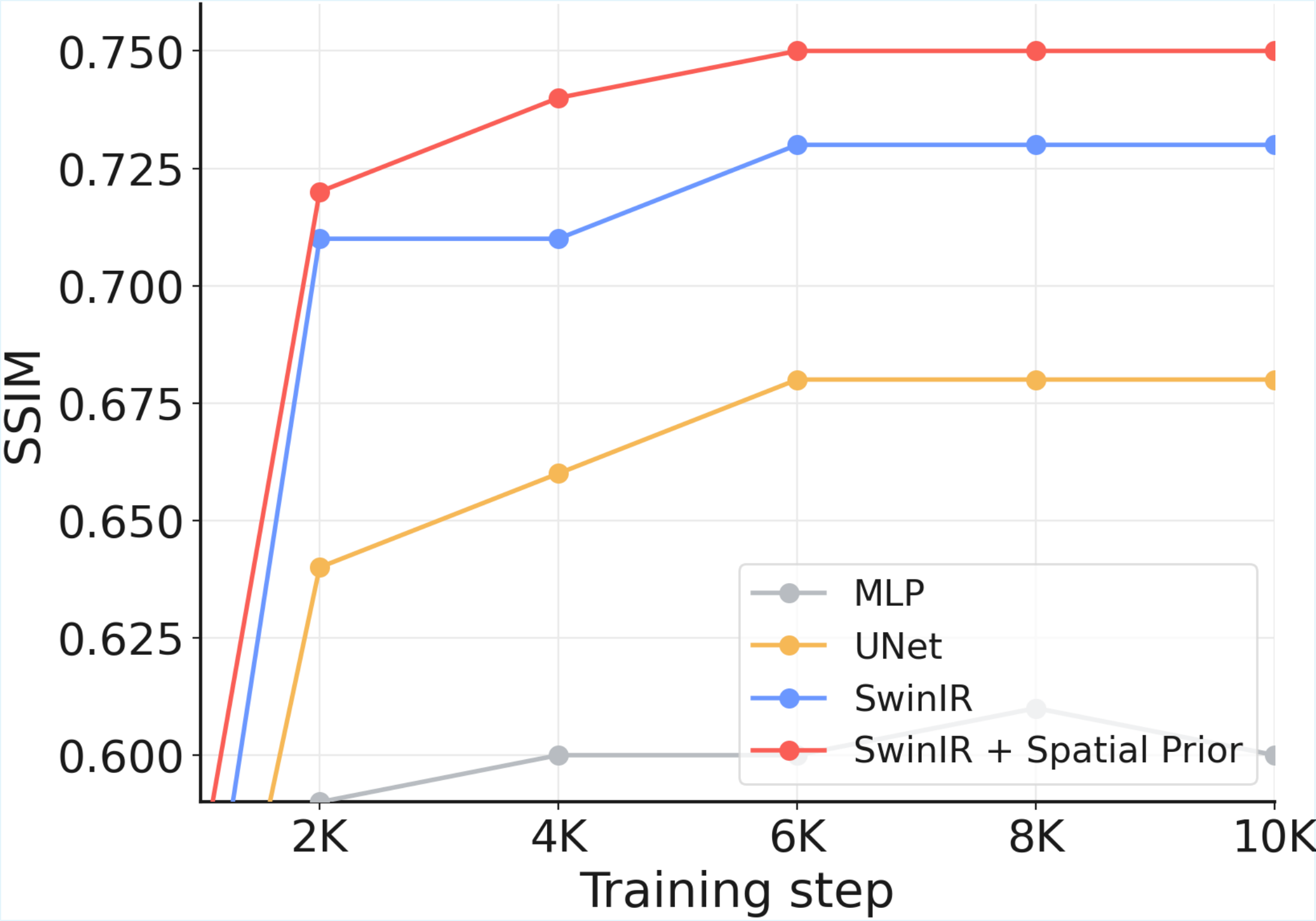}
{\scriptsize (e) Convergence across architectures.}
\end{minipage}
\caption{\textbf{Bridge design ablations and convergence.}
\textbf{Left}: Ablations of Bridge training, design components, objectives, and alignment architectures using decoded-image fidelity metrics. Attention Fidelity provides a large gain, and the full Bridge with Spatial Prior plus Attention Fidelity attains the best fidelity.
\textbf{Right}: SSIM over training iterations. Attention Fidelity speeds up convergence and increases the final SSIM. Under matched settings, the Spatial Prior alignment architecture converges to higher SSIM than alternatives. L1 reconstruction is used in all experiments.}
\label{fig:bridge_ablation}
\end{figure}

\cref{fig:bridge_ablation} summarizes our Bridge design ablations and convergence behavior. We report reconstruction metrics L1, PSNR, and SSIM in \emph{image space}, computed between the decoded outputs $\hat{x}=\mathcal{D}_T(\hat{z}_T)$ and $x=\mathcal{D}_T(z_T)$. All variants are trained on the same data with the same optimization schedule.

\myheading{Architectures.}
We compare three backbones: MLP, UNet, and SwinIR. For SwinIR, we additionally evaluate our frozen spatial prior, implemented as a fixed prefix of the Student decoder that expands $z_S$ to the Teacher grid before projection. The MLP underperforms, while UNet and SwinIR provide stronger spatial modeling. Adding the spatial prior to SwinIR gives a clear gain, indicating that reusing a pretrained upsampling scaffold helps the learnable projector focus on feature and semantic alignment rather than relearning spatial expansion.

\myheading{Objectives.}
We ablate objectives for learning $\Bridge_\phi$. Using only $\Loss_{rec}$ yields limited fidelity, and E-LatentLPIPS provides marginal improvement. In contrast, Attention Fidelity consistently enhances reconstruction and achieves the same SSIM in fewer steps, indicating that matching the Teacher denoiser's attention offers a stronger and more stable supervision in Teacher space.

\section{Limitations}
Our current study focuses on the most practically important form of heterogeneity in modern distillation pipelines: mismatch in latent resolution and VAE space. In this sense, Bridge is intentionally designed as a \emph{latent-space interface}: it aligns Student and Teacher representations after conditioning has been formed, while leaving the Student backbone unchanged. This design keeps the method lightweight, modular, and easy to integrate with existing one-step backbones, but it also means that differences in the conditioning stack, such as the text encoders used by modern foundation models (\eg, FLUX or SD~3.5) versus the CLIP-based conditioning used in compact students such as SD~1.5, are not explicitly modeled in the current formulation. Extending the same interface principle from latent alignment to conditioning-space alignment is a promising next step.

Our experiments also target a deliberate operating point: compact one-step image generation. Rather than reproducing the full capacity of large multi-step Teachers, the goal of this paper is to transfer as much Teacher knowledge as possible into efficient, ecosystem-compatible Students through a minimal additional module. Under this constraint, Bridge substantially narrows the gap to much larger Teachers while preserving the deployment advantages of compact backbones. We view this as a strength of the current formulation: it isolates the representation-alignment problem without requiring student redesign or large-scale retraining from scratch.

More broadly, we believe the scope of Cross-Space Distillation extends beyond the specific text-to-image setting studied here. Many modern generative systems operate over heterogeneous latent spaces, including latent video models, image editing pipelines, and other multimodal generators. The formulation introduced in this paper therefore suggests a broader research direction: treating alignment across heterogeneous latent representations as a reusable interface problem. Extending Bridge to video generation, editing models, and richer conditional pipelines is a natural next step, and one that could further increase the practical impact of compact generative models.

\section{Additional Qualitative Results}
\begin{figure}[!t]
    \centering
    \includegraphics[width=\linewidth]{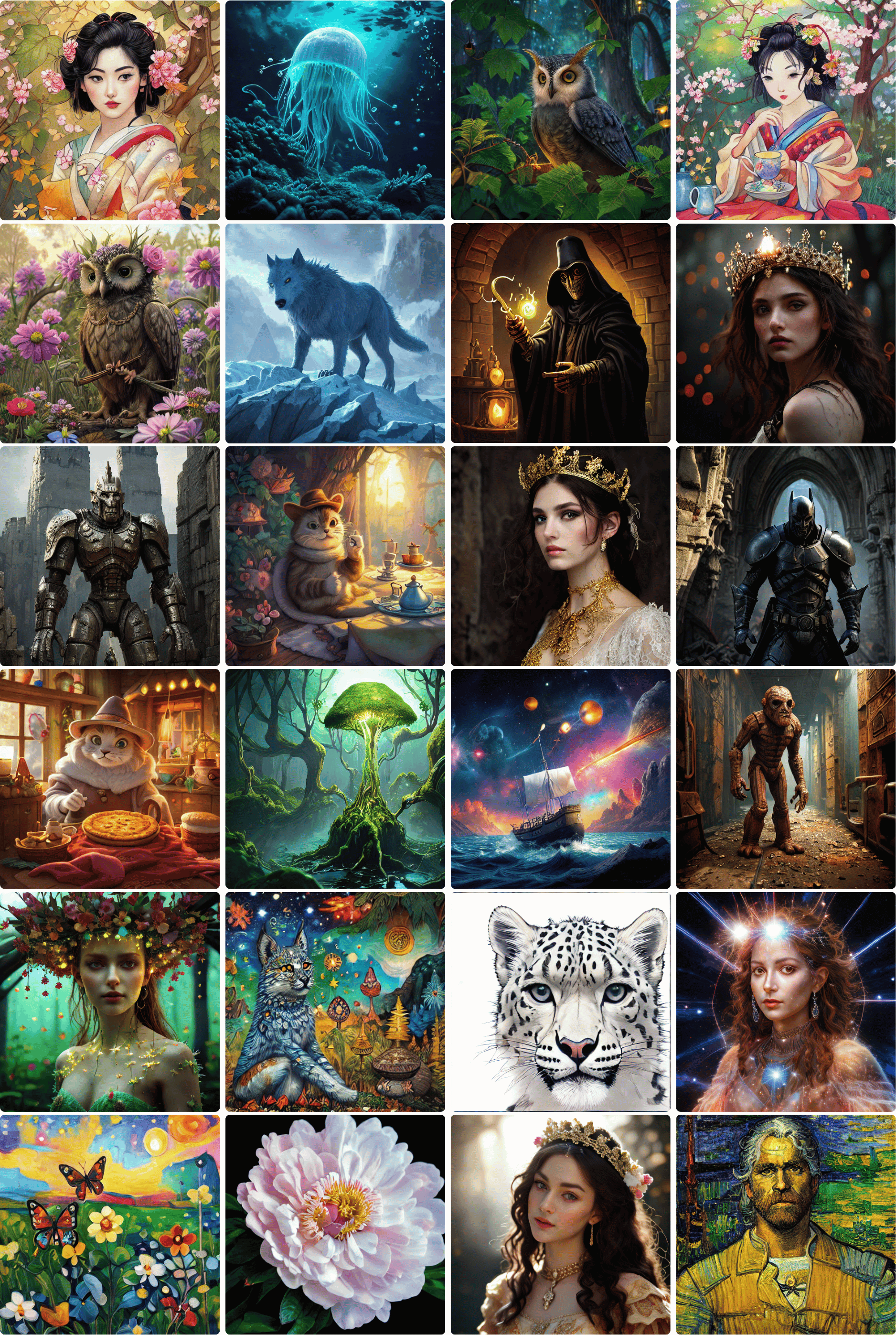}
    \caption{Additional Qualitative Results for our SD 1.5 Merged Model.}
    \label{fig:uncure_sd15}
\end{figure}

\begin{figure}[!t]
    \centering
    \includegraphics[width=\linewidth]{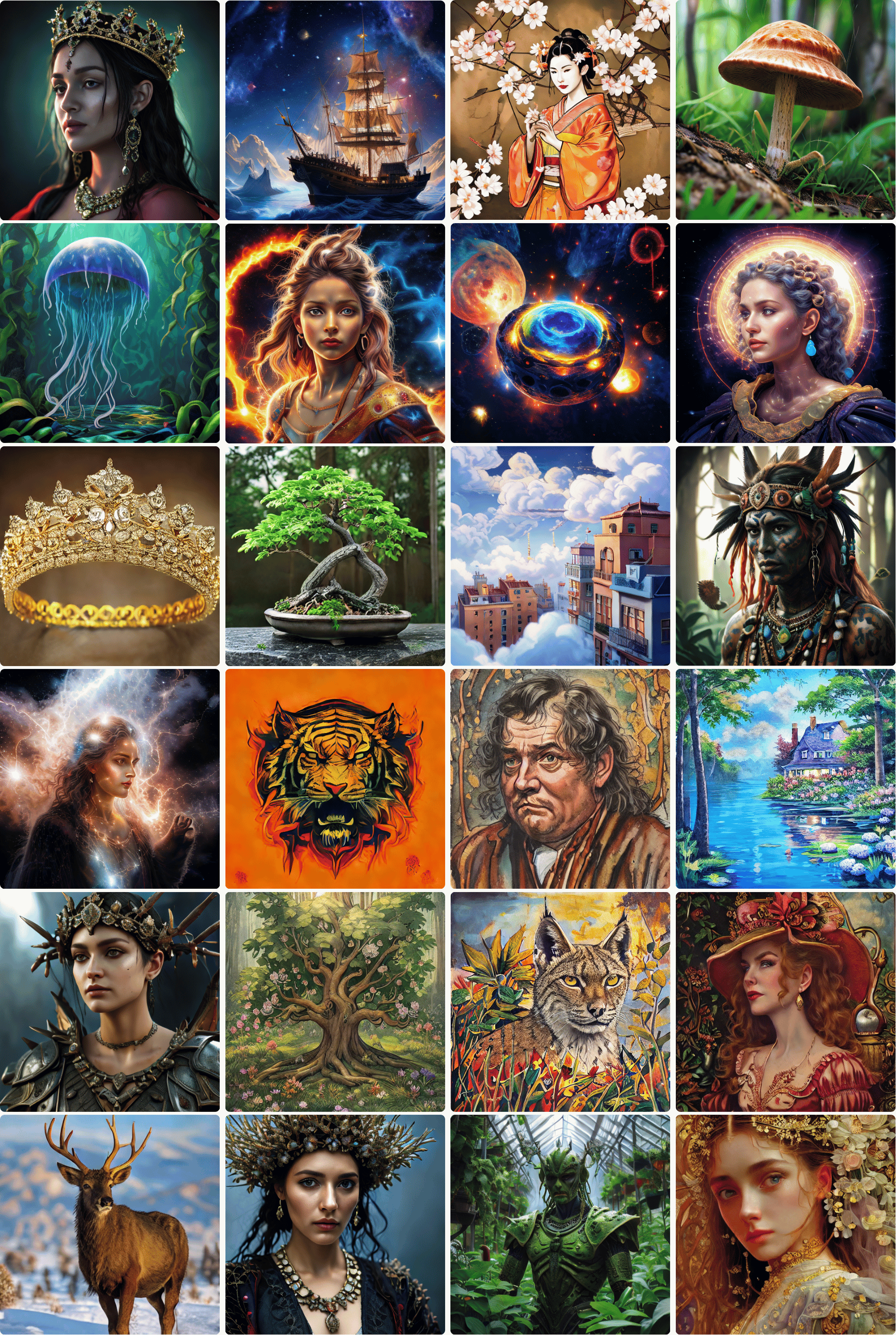}
    \caption{Additional Qualitative Results for our SD 2.1 Student Merged Model.}
    \label{fig:uncure_sd21}
\end{figure}

We provide additional uncurated samples generated by our distilled one-step SD 1.5 and SD 2.1 in \cref{fig:uncure_sd15} and \cref{fig:uncure_sd21}, respectively.

\section{Extended Related Work}
\subsection{Diffusion and Flow.}

Diffusion models have become the dominant framework for high-quality image generation. Early works such as DDPM \cite{ddpm} and score-based SDE models \cite{score_sde} established the standard formulation of learning a reverse denoising process from progressively corrupted data. Subsequent works improved the efficiency of this framework. DDIM \cite{ddim} introduced a deterministic sampling path that substantially reduces the number of denoising steps, while DPM-Solver \cite{dpm_solver} and EDM \cite{edm} further improved fast sampling through better numerical solvers, parameterization, and noise design.

In parallel, continuous-time transport formulations provide another important view of generative modeling. Flow Matching \cite{flow_matching} learns the velocity field of a probability path between noise and data, offering a simple alternative to diffusion-style objectives. Rectified Flow \cite{rectified_flow} further studies straighter transport paths to simplify generation trajectories, and OT-based conditional flow matching \cite{cfm_ot} improves path quality through optimal transport based couplings. These formulations have recently become increasingly relevant for large-scale image generation.

Modern text-to-image systems build on these advances with stronger backbones, larger training data, and higher-capacity latent representations. Stable Diffusion~3.5 \cite{sdv3} and Flux \cite{blackforest2024flux1,blackforest2025flux2} are representative examples that achieve strong visual fidelity and prompt alignment. However, these gains typically come with higher inference cost and larger model size, which makes direct deployment difficult in practical resource-constrained settings.

\subsection{Distillation for Fast Diffusion Sampling.}

To reduce inference cost, many works study distilling multi-step diffusion models into one-step or few-step generators. Progressive Distillation \cite{progressive_distillation} is an early and influential approach that progressively shortens the sampling trajectory while preserving the behavior of the original model. Consistency-based methods follow a related goal. Latent Consistency Models \cite{lcm} train a latent generator that supports high-quality few-step inference, and Truncated Consistency Models \cite{tcm} further improve efficiency by learning from shortened trajectories.

Another important line is distribution-based distillation. SwiftBrush \cite{vsd} and SwiftBrush v2 \cite{sb1} show that one-step generators can be trained effectively by matching Teacher and Student predictions on noised samples. Related methods such as one-step diffusion with $f$-divergence distribution matching \cite{f_distill}, SNOOPI \cite{snoopi}, and more recent distribution matching approaches \cite{yin2024improved,yin2024onestep} further improve this idea by directly optimizing the Student toward the Teacher-induced data distribution. These methods have significantly narrowed the quality gap between fast Students and their multi-step Teachers.

Several works also combine distillation with adversarial supervision. ADD \cite{add} introduces a discriminator to provide stronger perceptual guidance during one-step distillation, and LADD \cite{ladd} extends this idea to latent high-resolution image synthesis. Such methods are effective for improving sharpness and realism, especially when standard distillation losses alone are not sufficient. However, most of these approaches assume that Teacher and Student operate in the same latent space, or at least under compatible latent resolution and VAE parameterization. This assumption is natural for within-family distillation, but becomes restrictive when the Teacher and Student come from different model families.

\myheading{Efficient and Mobile Diffusion Models.}
Another line of research aims to reduce the physical size and computational footprint of diffusion models to enable deployment on resource-constrained devices. Approaches such as FastFlux \cite{fastflux}, SnapFusion \cite{snapfusion}, MobileDiffusion \cite{mobilediffusion}, and SnapGen \cite{snapgen,snapgen++} explore techniques including network pruning, structural compression, and lightweight architecture design. These methods demonstrate that diffusion models can be significantly compressed while maintaining acceptable image quality.

However, such approaches face several practical challenges. Network pruning often encounters a sparsity threshold beyond which further compression leads to noticeable quality degradation. Lightweight architectures typically require substantial redesign and retraining tailored to specific Teacher models.
% Moreover, customized architectures may break compatibility with the extensive ecosystem built around widely adopted backbones such as Stable Diffusion 1.5 \cite{sd1}, where extensions including ControlNet \cite{controlnet} and style adapters rely on fixed architectural configurations.

Our goal is different from these approaches. Instead of modifying the Student backbone itself, we keep the compact Student architecture unchanged and study how to transfer knowledge from a stronger Teacher even when the two models use different latent spaces. In this sense, our setting is complementary to efficient-model design and model compression.

\subsection{Representation Alignment and Internal Distillation.}

Our work is also related to distillation methods that align internal representations instead of only matching final outputs. In language modeling, MiniLLM \cite{guminillm} shows that reverse-KL based distillation can better preserve the dominant behavior of a strong Teacher. In diffusion models, X2I \cite{Ma_2025_ICCV} shows that attention distillation can serve as an effective supervision signal for transferring useful internal structure across models.

These observations are closely related to our design. Rather than treating the Student output alone as the target of distillation, we also consider how internal representations can be aligned when Teacher and Student are mismatched. Different from prior works, however, our focus is specifically on the case where the gap comes from both latent resolution and latent parameterization. Our Cross-Space Distillation addresses this problem with a lightweight Bridge that maps Student features into the Teacher latent space, so that standard distillation objectives can still be applied without changing the Student architecture.

\subsection{Comparison to Deep Compression Autoencoder Approaches.}
Our technique achieves efficient image generation by distilling from a big teacher operating at resolution $1024\times1024$ to a compact student operating at resolution $512\times512$. Some recent methods achieve similar efficiency using deep compression autoencoders, such as the SANA family \cite{xie2024sana,xie2025sana,chen2025sana}. However, our method offers several advantages: (1) \textbf{Efficient training.} SANA and SANA 1.5 need to be trained from scratch, using a large amount of training data and at least 100K training iterations. Even the distillation process to produce SANA-Sprint requires 25000 iterations on 32 A100 GPUs, using the same training data. In contrast, our method relies on score distillation to a student model that is well initialized with existing pretrained weights. Hence, our training is highly efficient, requiring only 2M synthetic images and around 10K training iterations. (2) \textbf{Simple architecture.} SANA models are based on the DiT structure. To achieve high inference speed, they need to employ advanced techniques such as Linear Attention and Triton-accelerated modules. Our method instead can reuse the standard UNet backbones in Stable Diffusion models, (3) \textbf{Onboard-friendliness.} Since SANA models are based on tailored architectures, onboarding them to devices requires significant effort. In contrast, our distilled students can be easily deployed on edge devices and are compatible with a broad range of existing applications.

\section{Societal Impacts}
Our work aims to enable a highly efficient framework for fast, accessible, and high-quality image generation. At the same time, we recognize that advanced image manipulation methods may be misused to create deceptive content \cite{inverfill,flexedit,efhq}. To address these concerns, we emphasize the importance of developing robust detection approaches for AI-generated or manipulated media \cite{anti-i2v, suma,erase,cgce}, alongside promoting the responsible deployment of such technologies.

\ifdefined\SUPPLEMENTARYINPUT
\clearpage
\else
\clearpage
\bibliographystyle{splncs04}
\bibliography{main}
\end{document}
\fi

\end{document}